\documentclass[sigconf]{acmart}

\usepackage[utf8]{inputenc} 
\usepackage[T1]{fontenc}    
\usepackage{hyperref}       
\usepackage{url}            
\usepackage{booktabs}       
\usepackage{amsfonts}       
\usepackage{nicefrac}       
\usepackage{microtype}      
\usepackage{xcolor}         
\usepackage{graphicx}
\usepackage{algorithm}
\usepackage{algorithmic}
\usepackage{color, colortbl}
\usepackage{array}
\usepackage{multirow}
\usepackage{enumitem}
\usepackage{amsmath}
\usepackage{arydshln}
\usepackage{rotating}
\usepackage{subfigure}
\usepackage{tabularx}
\usepackage{relsize}
\definecolor{Gray}{gray}{0.9}

\usepackage{newfloat}
\usepackage{listings}
\floatstyle{ruled}
\newfloat{listing}{tb}{lst}{}
\floatname{listing}{Listing}

\AtBeginDocument{%
  }

\copyrightyear{2024}
\acmYear{2024}
\setcopyright{acmlicensed}\acmConference[SIGSPATIAL '24]{The 32nd ACM International Conference on Advances in Geographic Information Systems}{October 29-November 1, 2024}{Atlanta, GA, USA}
\acmBooktitle{The 32nd ACM International Conference on Advances in Geographic Information Systems (SIGSPATIAL '24), October 29-November 1, 2024, Atlanta, GA, USA}
\acmDOI{10.1145/3678717.3691241}
\acmISBN{979-8-4007-1107-7/24/10}

\begin{document}
\pagestyle{plain}
\title{SAUC: Sparsity-Aware Uncertainty Calibration for Spatiotemporal Prediction with Graph Neural Networks}

\author{Dingyi Zhuang}
\affiliation{%
  \institution{Massachusetts Institute of Technology}
  \city{Cambridge}
  \country{USA}
}
\email{dingyi@mit.edu}

\author{Yuheng Bu}
\affiliation{%
  \institution{University of Florida}
  \city{Gainesville, Florida}
  \country{USA}
}
\email{buyuheng@ufl.edu}

\author{Guang Wang}
\affiliation{%
  \institution{Florida State University}
  \city{Tallahassee, Florida}
  \country{USA}
}
\email{guang@cs.fsu.edu}

\author{Shenhao Wang}
\affiliation{%
  \institution{University of Florida}
  \city{Gainesville, Florida}
  \country{USA}
}
\email{shenhaowang@ufl.edu}

\author{Jinhua Zhao}
\affiliation{%
  \institution{Massachusetts Institute of Technology}
  \city{Cambridge}
  \country{USA}
}
\email{jinhua@mit.edu}

\renewcommand{\shortauthors}{Zhuang et al.}

\begin{abstract}
  Quantifying uncertainty is crucial for robust and reliable predictions. 
  However, existing spatiotemporal deep learning mostly focuses on deterministic prediction, overlooking the inherent uncertainty in such prediction. 
  Particularly, highly-granular spatiotemporal datasets are often sparse, posing extra challenges in prediction and uncertainty quantification. 
  To address these issues, this paper introduces a novel post-hoc Sparsity-aware Uncertainty Calibration (SAUC) framework, which calibrates uncertainty in both zero and non-zero values. 
  To develop SAUC, we firstly modify the state-of-the-art deterministic spatiotemporal Graph Neural Networks (ST-GNNs) to probabilistic ones in the pre-calibration phase. 
  Then we calibrate the probabilistic ST-GNNs for zero and non-zero values using quantile approaches.
  Through extensive experiments, we demonstrate that SAUC can effectively fit the variance of sparse data and generalize across two real-world spatiotemporal datasets at various granularities. 
  Specifically, our empirical experiments show a 20\% reduction in calibration errors in zero entries on the sparse traffic accident and urban crime prediction. 
  Overall, this work demonstrates the theoretical and empirical values of the SAUC framework, thus bridging a significant gap between uncertainty quantification and spatiotemporal prediction. 
\end{abstract}



\maketitle

\section{Introduction}
Spatiotemporal Graph Neural Networks (ST-GNNs) have been instrumental in leveraging spatiotemporal data for various applications, from weather forecasting to urban planning \citep{wu2021inductive, zhao2022multi,gao2023spatiotemporal}. 
However, most of the models narrowly focus on deterministic predictions, which overlooks the inherent uncertainties associated with the spatiotemporal phenomena. 
For various safety-related tasks, such as traffic accidents and urban crime prediction, it is crucial to quantify their uncertainty, thus proactively preventing such phenomena from happening. 
Despite its importance, uncertainty quantification (UQ) remains relatively understudied in the spatiotemporal context, which can lead to erroneous predictions with severe social consequences \citep{chung2021beyond,abdo2017uncertainty,rudin2019stop}.

Recent research has addressed UQ in spatiotemporal modeling \citep{wang2023uncertainty,zhuang2022uncertainty,huang_uncertainty_2023}, but often fails to account for the sparsity and asymmetrical distribution present in detailed spatiotemporal data. 
This sparsity, marked by an abundance of zeros, becomes essential when working with high-resolution data. 
Additionally, the need for granular forecasting in event precaution often leads to more diluted inputs, making sparsity unavoidable. 
Thus, effectively managing aleatoric (data) uncertainty is crucial in these forecasting tasks.

Prior ST-GNNs with basic UQ functionalities typically focus on assuming data distributions (e.g., Gaussian) and learning these distributions' parameters for spatiotemporal predictions \citep{jiang2023uncertainty,zhuang2022uncertainty,wang2023uncertainty}. 
A key limitation is their inadequate evaluation of UQ reliability using robust calibration metrics. Research by \citep{zhuang2022uncertainty,wu2021quantifying,jiang2023uncertainty,gao2023spatiotemporal} often emphasizes high accuracy and narrow prediction intervals as indicators of effective UQ. 
This approach overlooks the crucial relationship between data variance and prediction intervals (PI), which is vital for addressing aleatoric uncertainty.
As highlighted by \citep{kuleshov2018accurate}, most deep learning models are not fully calibrated for high UQ quality, leaving a significant research gap to be filled.

This paper introduces a Sparsity-Aware Uncertainty Calibration (SAUC) framework, calibrating model outputs to align with true data distributions, which can be adapted to any ST-GNN model. 
Inspired by the zero-inflated model \citep{agarwal2002zero,fernandes2009modelling}, we partition zero and non-zero predictions and apply separate Quantile Regression (QR) models to ensure the discrepancy of prediction and true values lie within the PIs. 
This post-hoc approach, unlike uncertainty-aware spatiotemporal prediction models \citep{rahman_uncertainty-aware_2023}, offers flexibility in applying it to existing models without necessitating architectural changes or full model retraining. 
Consistent with prior studies on regression task calibration, we formulate a modified calibration metric tailored for PI-based calibration within asymmetric distributions. 
To investigate our model's generalizability, we adapt three prevalent numeric-output ST-GNN models with negative binomial (NB) distributions to accommodate sparse settings without compromising prediction accuracy. 
Tested on two real-world spatiotemporal datasets with different levels of sparsity, our SAUC demonstrates promising calibration results, thus assisting in reliable decision-making and risk assessment.
The main contributions of this paper are threefold:
\begin{itemize}
    \item We have developed a novel post-hoc uncertainty calibration framework tailored for predictions on sparse spatiotemporal data. 
    This approach is designed to be compatible with current GNN models and is suitable for any probabilistic outputs.
    \item We focus on calibrating asymmetric distributions, like the NB distributions, using quantile regression. 
    This approach deviates from conventional mean-variance UQ methods. 
    Additionally, we have developed new calibration metrics based on prediction interval width.
    \item We conduct experiments on real-world datasets with different temporal resolutions to demonstrate promising calibration results of our SAUC, especially for zero values.
    We observe a roughly \textbf{20\% reduction} in calibration errors relative to leading baseline models, underscoring its efficacy for safety-critical decision-making.
\end{itemize}

\section{Related Work}

\subsection{Spatiotemporal Prediction}
Spatiotemporal prediction through deep learning offers a powerful tool for various applications. Among them, GNNs have become prevalent in recent years \citep{wu2021inductive,zhao2022multi,yu2017spatio,wu2019graph}. 
Most existing models focus on predicting the deterministic values without considering the associated uncertainty, which leaves a noticeable research gap concerning UQ. 
Such a limitation impacts the reliability of predictions, as the unaccounted uncertainty can lead to deviations from the true data variance.

Bayesian and the frequentist approaches are the two main existing methods for spatiotemporal UQ \citep{wang2023uncertainty,zhou2021stuanet}. 
While the Bayesian approach offers profound insights through techniques like Laplace approximation \citep{wu2021bayesian}, Markov Chain Monte Carlo \citep{pearce2018high}, and Variational Inference \citep{khosravi2010lower}, it suffers from computational complexities and approximation inaccuracies. 
In contrast, the frequentist approach exploits Mean-Variance Estimation for computational efficiency and model flexibility, accommodating both parametric and non-parametric methods \citep{wang2023uncertainty}.
Recently, many frequentist approaches are proposed by modifying the last layers of the ST-GNN model outputs \citep{jiang2023uncertainty,gao2023spatiotemporal,rahman_uncertainty-aware_2023}.
However, the application of these methods to sparse data is challenged by data dependencies and distribution constraints \citep{wu2023adaptive,gawlikowski2021survey}. 
For example, sparse data frequently deviate from models' mean-variance assumptions, favoring other right-skewed distributions like NB over Gaussian, and traditional metrics like mean squared errors can be skewed by sparse datasets' non-zero values.

The use of zero-inflated distributions with spatiotemporal neural networks has been introduced to handle zero instances and non-normal distribution in sparse data \citep{zhuang2022uncertainty}. 
However, this approach has limitations in extreme scenarios and non-time-series contexts, impacting predictive accuracy and system robustness \citep{qian2023uncertainty}. 
Therefore, the urgent need to quantify and calibrate uncertainty in the sparse component of spatiotemporal data is evident, yet remains largely unexplored.

\subsection{Uncertainty Calibration}
Although relatively understudied in the spatiotemporal context, UQ has been examined by many machine learning studies. 
Among all the UQ approaches, calibration is a post-hoc method that aims to match the predictive probabilities of a model with the true probability of outcomes, developing robust mechanisms for uncertainty validation \citep{nixon2019measuring}.

Many techniques have been developed to calibrate pre-trained classifiers, mitigating the absence of inherent calibration in many uncertainty estimators \citep{kumar2019verified,gruber2022better,shen2023post}. 
Calibration for regression tasks, however, has received less attention \citep{thiagarajan2020building,kuleshov2018accurate}. 
Post-hoc calibration methods for classifications, such as temperature scaling, Platt scaling, and isotonic regression, have successfully been adapted for regression \citep{kull2019beyond}. 
Recent research \citep{chung2021beyond} further demonstrated quantile regression-based calibration methods and their extensions. 
These post-hoc methods offer greater flexibility and wider applicability to various probabilistic outputs compared to in-training methods, without necessitating model retraining \citep{rahman_uncertainty-aware_2023,huang_uncertainty_2023}.

In spatiotemporal prediction contexts, where accurate model outputs are important for applications like traffic accident forecasting and crime prediction \citep{zhao2022multi}, the significance of uncertainty calibration is often overlooked. 
Since spatial and temporal granularity of the data could vary, there could exist different levels of sparsity in the data \citep{zhuang2022uncertainty}.
It is essential to discern confidence for zero and non-zero segments due to their varied implications. 
In response, we propose the SAUC framework to manage asymmetric distributions and ensure variances of true targets align with estimated PIs \citep{thiagarajan2020building}.

\section{Problem Formulation}
\subsection{ST-GNNs for Prediction} 
\label{sec:stgnn}
\noindent Let $\mathcal{G} = (\mathcal{V},\mathcal{E},\mathbf{A})$ where $\mathcal{V}$ represents the set of nodes (locations), $\mathcal{E}$ the edge set, and $\mathbf{A}\in \mathbb{R}^{|\mathcal{V}|\times|\mathcal{V}|}$ the adjacency matrix describing the relationship between nodes. 
We denote the spatiotemporal dataset $\mathcal{X} \in \mathbb{R}^{|\mathcal{V}| \times t}$ where $t$ is the number of time steps. The objective is to predict the target value $\mathcal{Y}_{1:|\mathcal{V}|,t:t+k}$ of future $k$ time steps given all the past data up to time $t$, which is $\mathcal{X}_{1:|\mathcal{V}|,1:t}$. 
The full dataset is partitioned by timesteps into training, calibration (i.e., validation), and testing sets. 
The three data partitions are denoted as $\mathcal{X}_{T}$, $\mathcal{X}_{S}$, and $\mathcal{X}_{U}$, respectively. 
For example, the training set $\mathcal{X}_{1:|\mathcal{V}|,1:t}$ is denoted as $\mathcal{X}_{T}$. 
The associated target values are represented as $\mathcal{Y}_{T}$, $\mathcal{Y}_{S}$,  and $\mathcal{Y}_{U}$, with subscripts indicating the corresponding sets.

The objective of the ST-GNN models is to design model $f_{\theta}$, parametrized by $\theta$, which yields:
\begin{equation}
    \hat{\mathcal{Y}}_d = f_{\theta}(\mathcal{X}_d; \mathcal{G}), \forall\ d \in \{ T,S,U \},
    \label{eq:GNN}
\end{equation}
where $\hat{\mathcal{Y}}_d$ is the predicted target value, and the subscript represents the corresponding data set. Note that $\theta$ is fixed once $f_{\theta}$ is trained on the set $T$.

With emerging interests in quantifying the data and model uncertainty of ST-GNNs, researchers shift towards designing probabilistic GNNs to predict both mean values and PIs \citep{wang2023uncertainty,bi2023mm}.
Previous probabilistic ST-GNN studies often assume $\mathcal{X}$ and $\mathcal{Y}$ are independent and identically distributed random variables, i.e., denoted as $X_i$ and $Y_i$ to describe data distributions.
The last layer of the neural networks architectures is modified to estimate the parameters of the assumed distributions\citep{zhuang2022uncertainty, jiang2023uncertainty}. 
For this study, we use the NB distribution to discuss the sparse and discrete data distribution. 

The NB distribution of each predicted data point is characterized by shape parameter $\mu$ and dispersion parameter $\alpha$, and $\hat{\mu}_i \in \hat{M}_d$ denotes the mean value of the predicted distribution. 
Previous work has applied prediction layers with the corresponding likelihood loss after ST-GNN encoding to study the parameter sets of the predicted data points \citep{zhuang2022uncertainty,jiang2023uncertainty}. 
The set of predicted $\mu$ and $\alpha$ corresponding to all elements of $\mathcal{Y}$ are denoted as $\hat{M}$ and $\hat{H}$. Therefore, Equation \ref{eq:GNN} can be rewritten in the format of NB distribution-based probabilistic ST-GNNs as:
\begin{equation}
    (\hat{M}_d,\hat{H}_d) = f_{\theta} (\mathcal{X}_d; \mathcal{G}), \forall d \in \{ T,S,U \}.
    \label{eq:PGNN}
\end{equation}
Subsequently, the objective function is adjusted to incorporate the relevant log-likelihood, as denoted in Equation \ref{eq:modify_loss}.

It is important to note that the selection of the distribution is not the primary focus of this paper.
Our goal is to present a general framework to calibrate probabilistic outputs for sparse spatiotemporal data, as long as the predicted distributions are provided. 
Implementations for other distributions are included in Section \ref{sec:poisson}, but the NB distribution is used for illustration in this study.

\begin{figure*}[!t]
    \centering
    \includegraphics[width= 0.9\textwidth ]{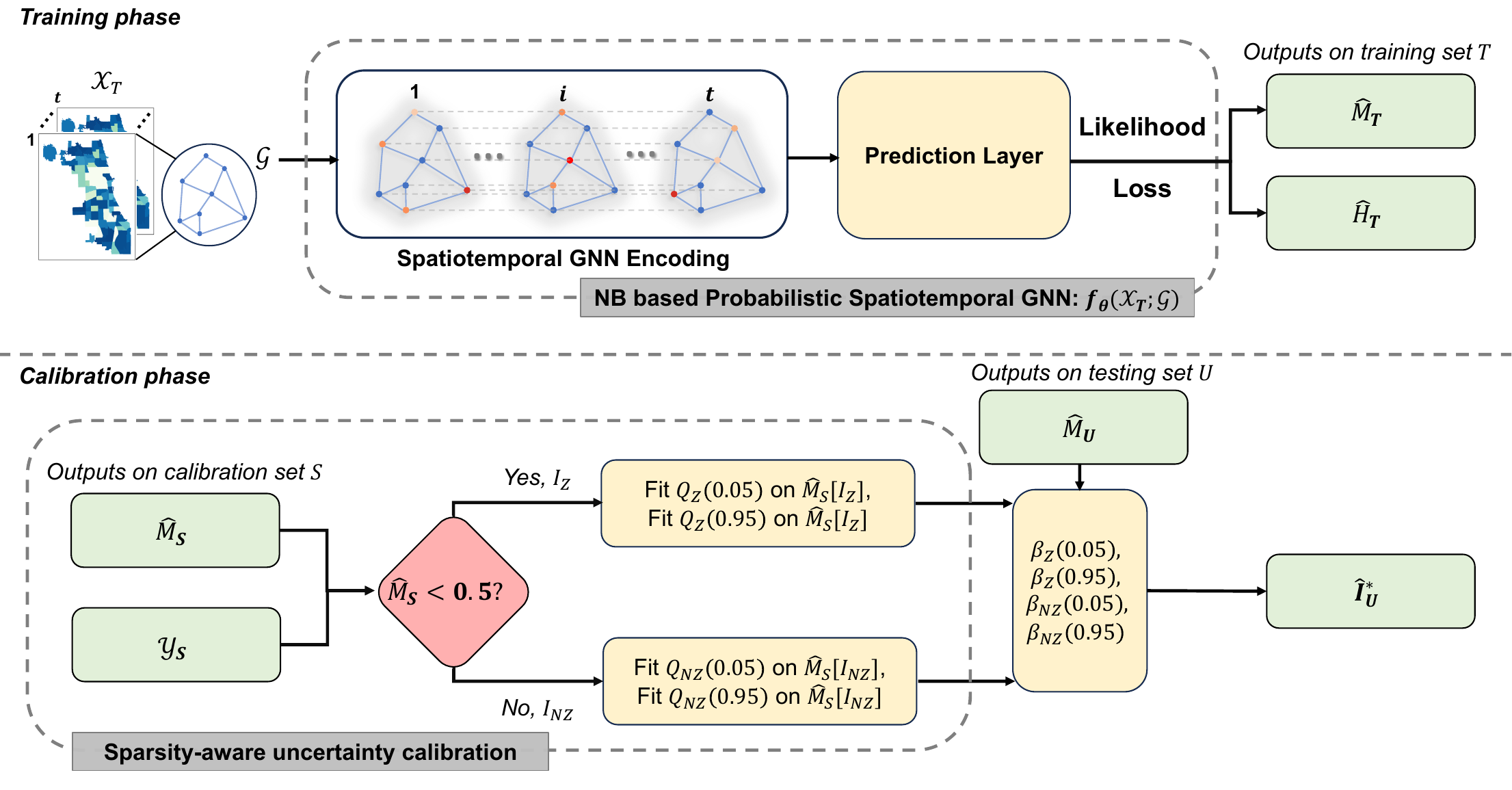}
    \caption{Experiments and modules of the SAUC framework. We modify existing ST-GNN models into probabilistic ones with NB distribution parameters as the outputs. Sparsity-aware uncertainty calibration is then conducted to obtain the quantile regression results and calibrated confidence interval for forecasting. These two steps can be easily and practically generalized to all existing ST-GNN models.}
    \label{fig:model}
\end{figure*}

\subsection{Prediction Interval Calibration}

Generally, for the frequentist approach, the spatiotemporal prediction models output the sets of location and shape parameters $\hat{M}_d,\hat{H}_d, \forall d \in \{ T, S, U\}$ \citep{song2019distribution,levi_evaluating_2020,kuleshov2018accurate}. 
We construct the prediction intervals of the $i$-th data point as $\hat{\mathcal{I}}_i = [\hat{l}_i,\hat{u}_i]$ with $\hat{l}_i$ and $\hat{u}_i$ denoting the 5th- and 95th-percentiles, respectively. 
Notably, our discussion limits the prediction intervals to the 5\% and 95\% range. 
Based on \citep{kuleshov2018accurate}, our target is to ensure a prediction model $f_{\theta}$ to be calibrated, i.e.,
\begin{equation}
    \frac{ \sum_{i=1}^{|U|} \mathbb{I}\{ y_{i} \leq \hat{F}_{i}^{-1}(p) \} }{|U|} \xrightarrow{|U| \rightarrow \infty} p,\  \forall p \in \{0.05, 0.95\},
    \label{eq:pi_calib}
\end{equation}
where $p$ stands for the targeting probability values, $\mathbb{I}$ denotes the indicator function, and $\hat{F}_{i}(\cdot)$ is the CDF of the NB distributions parameterized by $\hat{\mu}_{i}$ and $\hat{\alpha}_{i}$. 
The primary goal is to ensure that, as the dataset size increases, the predicted CDF converges to the true one. 
Such an alignment is congruent with quantile definitions, suggesting, for instance, that 95\% of the genuine data should fall beneath the 95\% percentile. 
Rather than striving for a perfect match across $\forall p \in [0,1]$, we mainly focus on calibrating the lower bound and upper bound.

Empirically, we leverage the calibration set $S$ to calibrate the uncertainty in data. 
The empirical CDF for a given value $p \in \hat{F}_i (y_i)$ is computed \citep{kuleshov2018accurate,levi2022evaluating}:
\begin{equation}
    \hat{\mathcal{P}}(p) = \frac{ |\{ y_{i} | \hat{F}_i(y_i) \leq p, i = 1, \dots, |S| \} | }{|S|}.
    \label{eq:cdf}
\end{equation}
The calibration process entails mapping a calibration function $\mathcal{C}$, such as isotonic regression or Platt scaling, to the point set $\{ (p,\hat{\mathcal{P}}(p)) \}_{i=1}^{|S|} $. 
Consequently, the composition $\mathcal{C} \circ \hat{F}$ ensures that $\hat{\mathcal{P}}(p) = p$ within the calibration dataset. 
This composition can then be applied to the test set to enhance the reliability of predictions.

\section{Sparsity-Aware Uncertainty Calibration}

\subsection{QR for Uncertainty Calibration}
\label{sec:qr}
Unlike traditional regression techniques that target the conditional mean, QR predicts specific quantiles, which are essential for calibrating PIs. 
Given $\hat{M}_S$ and $\mathcal{Y}_S$ from the calibration set $S$, the linear QR model can be expressed as:
\begin{equation}
    Q_{\mathcal{Y}_S | \hat{M}_S }(p) = \hat{M}_S^T \beta(p),
    \label{eq:QR}
\end{equation}
where $\beta(p)$ signifies the coefficient associated with quantile $p$. The goal here is to minimize the Pinball loss \citep{chung2021beyond}, which is defined as: 

\begin{equation}
L_{\text{pinball}}(y, \hat{y}, p) =
\begin{cases}
(p-1) \cdot (y - \hat{y}), & \text{if } y < \hat{y} \\
p \cdot (y - \hat{y}), & \text{if } y \geq \hat{y}
\end{cases}.
\end{equation}
In our context, we focus on the cases of $p=5\%$ and 95\%, and thus the quantile regressions output calibrated PIs correspond to $[Q(0.05),Q(0.95)]$.

While the NB distribution is frequently used for estimating sparse data uncertainty in previous work \citep{zhuang2022uncertainty,jiang2023uncertainty,sankararaman2013distribution,chen2021joint}, it can still be prone to misspecification, particularly when assumptions about data characteristics are less precise. 
In such instances, non-parametric (not directly modeling the distribution parameters) and distribution-agnostic (not reliant on specific distribution assumptions) methods like QR emerge as robust alternatives, compared to a direct scaling or fitting of the dispersion parameter \citep{laves2020well,bishop2006pattern}. 
QR's resilience to data assumptions helps mitigate potential errors in model specification, a critical advantage especially in handling datasets with heavy tails where significant risks or events are concentrated—features sometimes not fully captured by parametric models \citep{hao2007quantile}. 

QR is also adept at addressing heteroscedasticity, where variance shifts with the response variable, a situation often oversimplified by a single dispersion parameter in parametric calibration methods \citep{zheng2018hybrid,antunes2017review}. 
Its ability to handle sparse datasets and accommodate variance fluctuations, particularly in scenarios with numerous zeros, further underscores its utility. 
Enhanced with basis expansion techniques, QR can discern complex patterns across various quantiles, making it an ideal choice for scenarios where zero outcomes are prevalent and a nuanced understanding of data patterns and risks is crucial.

\subsection{SAUC Framework}
Our SAUC framework is a two-step post-hoc calibration framework tailored for sparse data, as detailed in Figure \ref{fig:model}. 
In the training phase, we trained the modified NB-based probabilistic ST-GNNs to obtain the model parameters $\theta$ (modifications detailed in Section \ref{sec:model_modification}). 
In the calibration phase, this training yields $\hat{M}_S,\hat{H}_S$ on the calibration set for uncertainty calibration.
We partition the calibration set into $N$ bins based on the granularity of data variance we are interested in.
We set $N=15$ by default, leaving its selection discussed in Section \ref{sec:num_bins}.
Inspired by zero-inflated models, predictions in the calibration set $S$ are bifurcated based on indices: $I_{Z} = \{\hat{M}_S < 0.5\}$ for prediction values nearing zero, and $I_{NZ} = \{\hat{M}_S \geq 0.5\}$ for non-zero predictions. 
The value of 0.5 is a natural threshold for rounding operations to either 0 or 1.
Following this partitioning, two QR models, $Q_Z$ and $Q_{NZ}$, are trained separately, calibrating quantiles $p = \{ .05, .95 \}$ to fine-tune the mean and PI predictions. 
The calibrated model is then applied to predict model outputs on the testing set. 
The pseudocode is provided in Algorithm \ref{alg:sauc}, and its implementation can be found in the anonymous GitHub \footnote{\url{https://github.com/AnonymousSAUC/SAUC}}.

The SAUC framework focuses on calibrating PIs rather than modifying distribution parameters, aligning with the principles of previous post-hoc calibration methods like isotonic regression and Platt scaling \citep{niculescu2005predicting,kuleshov2018accurate}. 
These methods are designed to refine a model's outputs without directly altering internal parameters such as the parameters $\mu$ and $\alpha$ of NB distributions. 
This strategy ensures a distinct separation between modeling and calibration, enhancing flexibility and broad applicability. 
It avoids complicating the calibration process and potentially impacting the core modeling objectives. 
Additionally, post-hoc methods have demonstrated practical effectiveness and computational efficiency in improving the probabilistic accuracy of predictions, without necessitating retraining or modifications to the original model.

\begin{algorithm}[t]
\caption{SAUC Framework}
\begin{algorithmic}[1]
\REQUIRE $\hat{M}_S$, $\mathcal{Y}_S$, $N$, $\hat{M}_U$, $p\leftarrow \{ .05, .95 \}$
\STATE Define the bin thresholds $\mathcal{T}_S$ using percentiles of $\mathcal{Y}_S$.
\STATE Initialize $U^*$, $L^*$ in the same size as $\hat{M}_U$.
\FOR{each bin $i$ in $[1, N]$}
\STATE Extract indices $I$ of instances in bin $i$ based on $\mathcal{T}_S$.
\STATE Split $I$ to $I_{NZ}$ for $\hat{M}_S\geq 0.5$ and $I_Z$ for $\hat{M}_S<0.5$.
\FOR{each set $i$ in $\{I_{NZ}, I_Z\}$}
\FOR{each value $q$ in $p$}
\STATE Fit $Q(q)$ on $\hat{M}_S[i]$ and $\mathcal{Y}_S[i]$ based on Equation \ref{eq:QR} and obtain $\beta (q)$.
\ENDFOR
\STATE 
$L^*[i]\leftarrow \hat{M}_U^T \beta(.05)$, $U^*[i]\leftarrow \hat{M}_U^T \beta(.95)$
\ENDFOR
\ENDFOR
\STATE $\mathcal{I}^* \leftarrow [L^*, U^*] $.
\RETURN $\mathcal{I}^*$.
\end{algorithmic}
\label{alg:sauc}
\end{algorithm}


\citet{song2019distribution} argued that quantile regressions might not retain true moments for the targets with similar predicted moments because of the global averaging effects. 
Our SAUC framework addresses this limitation because it creates local segments based on zero and non-zero values. 
The zero segment, potentially dominant, requires precision in calibration due to its binary implications in real-world applications. 
For the non-zero subset, the aim is to ensure that quantiles mirror actual event frequencies, resonating with the typical research focus on discerning relative risks or magnitudes of events.

\subsection{Calibration Metrics for Asymmetric Distributions}
\subsubsection{Expected Normalized Calibration Error}
\begin{figure}[ht]
    \centering    \includegraphics[width=0.8\linewidth]{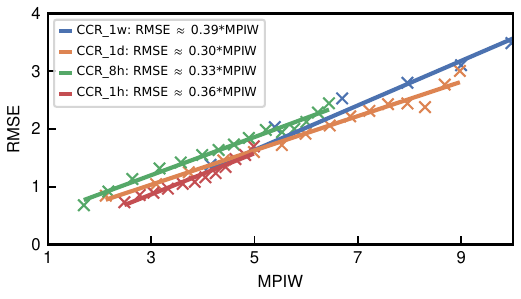}
    \caption{Explore the linear correlation between MPIW and RMSE using different cases of the CCR data.}    
    \label{fig:find_c}
\end{figure}

While calibration errors are clearly established for classification tasks using Expected Calibration Error (ECE) \citep{nixon2019measuring,gruber2022better,kumar2019verified}, the definition is much less studied in the regression context, where prediction errors and intervals supersede accuracy and confidence. 
Based on Equation~\ref{eq:pi_calib} and~\ref{eq:cdf}, we aim to calibrate the length of the prediction interval with the standard error, i.e.,
\begin{equation}
    \big[\mathbb{E}[ (\hat{\mu}_i - Y_i)^2 | \hat{\mathcal{I}_i} = \mathcal{I}_i ]\big]^{1/2} = c|\mathcal{I}_i|, \ \forall i \in \{ 1, \dots, |U| \}.
    \label{eq:define_calib}
\end{equation}
Note that $\mathcal{I}_i= [F_i^{-1}(.05),F_i^{-1}(.95)]$ is the realization of the predicted 5\%-95\% confidence interval and $|\mathcal{I}_i|$ denotes its width. 
Calibrating the model's standard errors with the 90\% confidence interval width reveals a linear relationship with coefficient $c$, specifically for symmetric distributions like Gaussian. 
As an approximation, we continue to use the z-score of Gaussian distribution to compute the coefficient $c$. 
The width of the 5\%-95\% confidence interval is roughly $2\times 1.645 = 3.29$ of the standard error, inversely proportional to $c$. 
Empirical results, illustrated in Figure \ref{fig:find_c}, support a linear correlation between the ratio of the predicted interval $\hat{\mathcal{I}}$ and the estimated standard error of the distribution. In our experiments, this slope ranges from 0.3 to 0.4, aligning closely with our theoretical assumption. Consequently, this leads to a value of $c \approx \frac{1}{3.29} \approx 0.303$.

Therefore, we could diagnose the calibration performance by comparing value differences on both sides of the equation. 
Similar to Algorithm \ref{alg:sauc}, we partition the indices of test set samples into $N$ evenly sized bins based on sorted predicted PI width, denoted as $\{B_j\}_{j=1}^N$. 
Each bin corresponds to a PI width that falls within the intervals: $[\min_{i\in B_j}\{ |\mathcal{I}_i| \}, \max_{i\in B_j}\{ |\mathcal{I}_i| \}]$. 
Note that the intervals are non-overlapping and their boundary values are increasing. 
Based on the discussion of \citet{levi2022evaluating} and Equation \ref{eq:define_calib}, for $j$-th bin, we define the Expected Normalized Calibration Error (ENCE) based on Root Mean Squared Error (RMSE) and Mean Prediction Interval Width (MPIW):
\begin{equation}
ENCE = \frac{1}{N} \sum_{j=1}^N \frac{|c \cdot MPIW(j) -  RMSE(j) |}{c \cdot MPIW(j)},
\label{eq:ence}
\end{equation}
where the $RMSE$ and $MPIW$ are defined as:
\begin{align}
    RMSE(j) &= \sqrt{ \frac{1}{|B_j|} \sum_{i\in B_j} (\hat{\mu}_i - y_i)^2 },\\
    MPIW(j) &= \frac{1}{|B_j|} \sum_{i \in B_j} |\hat{\mathcal{I}}_i|.
\end{align}
The proposed ENCE normalizes calibration errors across bins, sorted and divided by $c \cdot MPIW$, and is particularly beneficial when comparing different temporal resolutions and output magnitudes across datasets. 
The $c \cdot MPIW$ and observed $RMSE$ per bin should be approximately equal in a calibrated forecaster.

\subsubsection{Coverage}
In uncertainty calibration, coverage directly assesses the alignment between predicted uncertainty intervals and the true variability of the data. 
The coverage is calculated as how many points fall within the confidence interval, written as:
\begin{equation}
Coverage_{0.05-0.95} = \frac{1}{|U|} \sum_{i \in U} 1_{(y_i \geq \hat{F}_i^{-1}(0.05) \, \& \, y_i \leq \hat{F}_i^{-1}(0.95))},
\label{eq:coverage}
\end{equation}
where $\hat{F}_i^{-1}(0.05)$ and $\hat{F}_i^{-1}(0.95)$ are the predicted lower and upper confidence interval bounds, respectively. 
The expected coverage should be 0.9 for an ideal calibration.
A coverage value closer to the ideal value indicates that the predicted intervals encompass a large proportion of the observed data points, signifying that the uncertainty estimates are accurate and reliable. 

Both ENCE and coverage evaluate probabilistic model performance but focus on different aspects:
1) Coverage is calculated at the point distribution level, while ENCE operates at the binning level, allowing detailed analysis across different ranges of predicted uncertainty;
2) Coverage measures the proportion of observed outcomes within PIs, assessing the model’s ability to capture data variability directly. 
Meanwhile, ENCE evaluates the deviation between expected and actual calibration, providing a normalized error metric for accuracy and sharpness. 
As an extension of the ECE in previous calibration work \citep{thiagarajan2020building,kumar2019verified,gruber2022better,shen2023post,kuleshov2018accurate}, ENCE could adapt to distributions for sparse data like NB distributions while keeping the theoretical meaning of calibration error. 

\section{Experiments}

\subsection{Data Description}
Two spatiotemporal datasets are used: (1) Chicago Traffic Crash Data (CTC)\footnote{\url{https://data.cityofchicago.org/Transportation/Traffic-Crashes-Crashes/85ca-t3if}} sourced from 277 police beats between January 1, 2016 and January 1, 2023. 
The CTC data records show information about each traffic crash on city streets within the City of Chicago limits and under the jurisdiction of Chicago Police Department; 
(2) Chicago Crime Records (CCR)\footnote{\url{https://data.cityofchicago.org/Public-Safety/Crimes-2001-to-Present/ijzp-q8t2}} derived from 77 census tracts spanning January 1, 2003 to January 1, 2023. 
This dataset reflects reported incidents of crime (with the exception of murders where data exists for each victim) that occurred in the City of Chicago.
Despite both CTC and CCR data originating from the Chicago area, their disparate reporting sources lead to different spatial units: census tracts for CCR and police beats for CTC. 
Both datasets use the first 60\% timesteps for training, 20\% for calibration and validation, and 20\% for testing. 
Full data descriptions can be found in Table \ref{tab:appendix_data}.
\begin{table}[h]
    \small
    \centering
    \begin{tabular}{cc|cccc}
    \toprule
        Dataset & Resolution & Size & Sparsity &  Mean & Max\\
    \midrule              
       \multirow{4}{*}{CCR}  & \cellcolor{Gray}1-hour & \cellcolor{Gray}(77, 175321)& \cellcolor{Gray}67\% & \cellcolor{Gray}0.3 & \cellcolor{Gray}59\\
       & \cellcolor{Gray}8-hour & \cellcolor{Gray}(77, 21916) & \cellcolor{Gray}18\% & \cellcolor{Gray}2.7 & \cellcolor{Gray}206\\
       & 1-day  & (77, 7306)  & 4\%  & 8 & 223 \\ 
       & 1-week & (77, 1044)& $<0.1\%$ & 55.8 & 539\\
    \midrule
       \multirow{4}{*}{CTC}  & \cellcolor{Gray}1-hour & \cellcolor{Gray}(277, 61369) & \cellcolor{Gray}96\% & \cellcolor{Gray}$<0.1\%$ & \cellcolor{Gray}5 \\   
       & \cellcolor{Gray}8-hour & \cellcolor{Gray}(277, 7672) & \cellcolor{Gray}76\% & \cellcolor{Gray}0.4 & \cellcolor{Gray}9 \\
       & 1-day  & (277, 2558) & 47\% & 1.1 & 15 \\
       & 1-week & (277, 366) & 7\% & 7.3 & 45\\      
    \bottomrule
    \end{tabular}
    \caption{Characteristics of various datasets showing variation in sparsity at different temporal resolutions. Dataset sizes are represented as (spatial, temporal) dimension pairs. Sparse cases are marked grey.}
    \label{tab:appendix_data}
\end{table}

The temporal resolutions of the datasets are varied to demonstrate the ubiquitous sparsity issue and its practical significance in spatiotemporal analysis. 
Four temporal resolutions, 1 hour, 8 hours, 1 day, and 1 week are created for CTC and CCR datasets.
We designate the 1-hour and 8-hour cases of Crash and Crime datasets as sparse instances, due to a higher prevalence of zeros.

\begin{figure}[ht]
    \centering
    \includegraphics[width=0.9\linewidth]{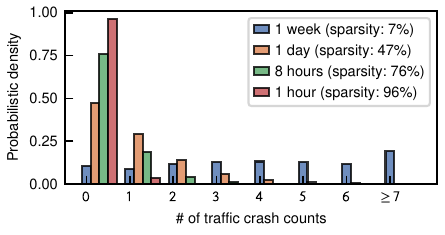}
    \caption{Histogram of number traffic crashes in the CTC data in different temporal resolutions}
   \label{fig:sparsity_transition}
\end{figure}

The histogram in Figure \ref{fig:sparsity_transition} shows traffic crash counts in all police beats across different temporal resolutions. 
Notably, higher resolutions show noticeable skewness and clustering of values at lower counts, illustrating the sporadic nature of traffic incidents. 

In the spatiotemporal data analysis, when we focus on more granular and responsive modeling, higher spatial or temporal resolutions are unavoidable.
This observation emphasizes the importance of equipping predictive models with capabilities to adeptly manage sparse data, thereby more accurately mirroring real-world temporal patterns.

For adjacency matrices, we calculate geographical distances $d_{ij}$ between the centroids of regions $i$ and $j$, representing census tracts or police beats. 
This distance is then transformed into a similarity measure, $\mathbf{A}_{ij} = e^{-d_{ij}/0.1}$, where $0.1$ is a scaling parameter, thus forming an adjacency matrix \citep{yu2017spatio}.

\subsection{Modified Spatiotemporal Models}
\label{sec:model_modification}
To demonstrate our model’s ability to generalize, we modify three popular ST-GNN forecasting models: Spatio-temporal Graph Convolution Network (STGCN) \citep{yu2017spatio}, Graph WaveNet (GWN) \citep{wu2019graph}, and Dynamic Spatial-Temporal Aware Graph Neural Network (DSTAGNN) \citep{lan2022dstagnn}. 
These models, originally designed for numerical outputs, are adapted to output the location parameter $\mu$ and dispersion parameter $\alpha$ of NB distributions. 
We add two linear layers to the neural network structures to learn these parameters separately, renaming the models STGCN-NB, GWN-NB, and DSTAGNN-NB, respectively. 
While NB distributions are used as an example, the output parameters are free to be adjusted based on data assumptions. 
Lastly, we replace the mean square error loss with the likelihood loss defined in Equation \ref{eq:modify_loss}:
\begin{equation}
\begin{aligned}
& \mathcal{L}(\mu, \alpha, y) = -\left[y \cdot \log\left(\frac{\mu + \epsilon}{\mu + \alpha + 2\epsilon}\right)\right] \\
&\quad\quad - \Gamma(y + \alpha + \epsilon) + \Gamma(y + 1) + \Gamma(\alpha + \epsilon) \\
&\quad\quad - \alpha \cdot \log\left(\frac{\alpha + \epsilon}{\mu + \alpha + 2\epsilon}\right) + \lambda \cdot ||\alpha||^2
\label{eq:modify_loss}
\end{aligned},
\end{equation}
where $y$ is the target variable, $\mu$ and $\alpha$ are model outputs, $\Gamma$ is the gamma function, $\lambda$ is the regularization parameter, and $\epsilon$ is a small constant added to improve numerical stability. 
Equation \ref{eq:modify_loss} is derived from the NB likelihood controlled by $\mu$ and $\alpha$.
Similar likelihood loss could be derived when choosing other distributions.

The loss function ensures the predicted NB distribution aligns well with the true target values. 
It penalizes predictions where the mean of the distribution ($\mu$) is far from the actual target values. 
Furthermore, it incorporates the shape of the distribution via the dispersion parameter ($\alpha$) to capture the variability in the data. 
The loss function also includes a regularization term to prevent overfitting by discouraging overly large values of $\alpha$.

All our experiments are implemented on a machine with Ubuntu 22.04, with Intel(R) Core(TM) i9-10980XE CPU @ 3.00GHz CPU, 128GB RAM, and NVIDIA GeForce RTX 4080 GPU.

\begin{table}[!h]
    \small
    \centering
    \begin{tabular}{cccc}
    \toprule
        Dataset & STGCN & GWN & DSTAGNN \\
    \midrule              
        CCR\_1h & 0.637 (0.703) & 0.962 (0.766) & 1.226 (1.117)\\
        CCR\_8h & 2.386 (2.326) & 2.052 (2.376) & 2.272 (2.693)\\
        CTC\_1h & 0.214 (0.220) & 0.980 (0.220) & 1.074 (0.871)\\
        CTC\_8h & 0.633 (0.726) & 0.924 (0.746) & 1.289 (1.177)\\
    \bottomrule
    \end{tabular}
    \caption{RMSE of original model outputs and the modified models (shown in the parenthesis) using sparse datasets.}
    \label{tab:model_modification}
\end{table}

As shown in Table \ref{tab:model_modification}, the model modification is successful without losing accuracy. 
In fact, the NB modification performs better in the sparse dataset as the NB distributions are more suitable for the nature of discrete data. 
Full implementation details and comparison of the modified ST-GNN models can be found in Appendix \ref{sec:modify_model_appendix}.


\setlength{\dbltextfloatsep}{1pt} %
\begin{table*}[!ht]
    \small
    \centering 
    \begin{tabularx}{\linewidth}{c*{7}{>{\centering\arraybackslash}X}@{\hspace{4pt}}c@{\hspace{4pt}}*{7}{>{\centering\arraybackslash}X}}
    \toprule
    {} & \multicolumn{7}{c}{Full observations} &  & \multicolumn{7}{c}{Zero-only targets} \\ 
    \midrule
    \smaller Method / & \smaller Before & \smaller Histogram & \smaller Isotonic & \smaller Temp. & \smaller Platt & \smaller QR & \smaller SAUC & \vline & \smaller Before & \smaller Histogram & \smaller Isotonic & \smaller Temp. & \smaller Platt & \smaller QR & \smaller SAUC \\
    \smaller Dataset    &  \smaller calibration    & \smaller  binning         &   \smaller regression       &  \smaller  scaling   &  \smaller  scaling &     &             &    \vline & \smaller calibration        &    \smaller  binning     &     \smaller regression     &         \smaller scaling  &  \smaller  scaling    &   &         \\
    \midrule
    \multicolumn{16}{c}{\textbf{STGCN-NB Outputs}} \\   
    \midrule
    CCR\_1h       &     1.160 &     0.410 &    0.517 &    0.887     &  0.443  &   \underline{0.235}  & \textbf{0.198}  &    \vline  &   1.167 &     0.361 &  0.386   &   0.473   &  0.477 &  \underline{0.273} &  \textbf{0.267}  \\
    CCR\_8h       &   1.389 &   \underline{0.344}   &    0.417 &    0.385  &  0.373 &  0.392 &  \textbf{0.336}    &  \vline  &   1.065 &     0.568 &    0.578 &       0.427 &  \underline{0.162} &  0.175  & \textbf{0.127}  \\
    CCR\_1d       &    0.413 &     0.593 &    0.546 &       0.364 &  \underline{0.294} &  0.297  &  \textbf{0.192}   &  \vline  &  1.266 &     1.047 &    0.979 &       1.167 &  \underline{0.840}  &    0.843    &  \textbf{0.738} \\
    CCR\_1w       &   0.855 &     0.784 &    0.773 &       0.765 &  0.579 & \underline{0.272} &    \textbf{0.242}   &    \vline  &   0.354 &     0.362 &    0.332 &       0.287 &  0.345 &  \underline{0.155}  &  \textbf{0.113}  \\    
    \midrule
    CTC\_1h       &    0.578 &     0.227 &    \underline{0.200} &       1.390 &  0.452 &  0.388 &   \textbf{0.165}  &       \vline  &   3.211 &     0.415 &    \underline{0.363} &       2.042 & 0.646 &  0.416  &   \textbf{0.078} \\
    CTC\_8h       &   0.323 &     0.320 &    0.255 &       0.570 &  0.297 &  \underline{0.241}  &  \textbf{0.233}   &       \vline  &  0.330 &     0.247 &    0.276 &       0.551 &  0.230 &  \underline{0.196}  &  \textbf{0.184} \\
    CTC\_1d       &   0.722 &     0.050 &     \textbf{0.011}  &       0.862 &  0.060 &  0.048  &  \underline{0.047} &       \vline  &  0.434 &     0.128 &    \textbf{0.027} &       5.451 &  0.152 &  0.379  & \underline{0.150}  \\
    CTC\_1w       &     2.279 &     1.227 &    1.527 &       0.914 &  2.265 &  \underline{0.378}   & \textbf{0.365} &       \vline  &  1.823 &     2.384 &    2.009 &       1.160 &  1.343 &  \underline{0.767}  &  \textbf{0.755}  \\

    \midrule
    \multicolumn{16}{c}{\textbf{GWN-NB Outputs}} \\
    \midrule
    CCR\_1h       &  1.200 &     1.022 &   0.889   &   1.149   &  0.834 & \underline{0.604}  & \textbf{0.493}   &   \vline  &  2.042 &     1.796 &    1.274 &       1.710 &  1.356 &  \underline{0.476}  &  \textbf{0.375} \\
    CCR\_8h       &    0.670 &     0.636 &    0.612 &   0.696 &  0.655 &  \underline{0.597} & \textbf{0.566}   &       \vline  &  1.398 &     0.389 &    0.492 &    1.179    &  0.478 &  \underline{0.308}   &  \textbf{0.186}   \\
    CCR\_1d       &   0.997  &     0.662 &    0.633 &    0.581 &  \underline{0.562} & 0.587  &  \textbf{0.504} &       \vline  &  1.158 &     0.583 &    \underline{0.580} &       0.725 &  0.747 &  0.644  & \textbf{0.563}  \\
    CCR\_1w       &    0.858 &     0.938 &    0.935 &       \textbf{0.611} &  0.931 &  0.857 &  \underline{0.829}  &       \vline  &  0.335 &     0.218 &    0.219 &       \textbf{0.161} &  0.219 & 0.250  &  \underline{0.172}    \\
    
    \midrule
    CTC\_1h      &   0.868  &  0.816  &  0.891  &  0.859  & \underline{0.634}  &  0.832  &  \textbf{0.290}  &  \vline  & 1.300  &  0.712  &  0.561  & 1.001  &  \underline{0.143}  &  0.998  & \textbf{0.139}  \\
    CTC\_8h      & 0.931 & 0.416 &  0.388 & 1.381 &  \underline{0.169} & 0.226  &\textbf{0.148} &        \vline & 2.213 &     0.720 &    0.659 &       1.995 &  \underline{0.421} &  0.776   &\textbf{0.221}\\
    CTC\_1d      & 0.523 & \underline{0.420} & 0.448 & 0.571 &  0.523 & 0.432   &\textbf{0.139}  &        \vline & 0.250 &     0.464 &    0.405 &       \underline{0.248} &  0.484 &   0.439  &\textbf{0.210} \\
    CTC\_1w      & 0.408 & 0.115 & \textbf{0.114} & 0.259 &  \underline{0.236} & 0.241 &0.239 &        \vline & 0.466 &     0.479 &    \textbf{0.436} &       0.457 &  0.451 &  0.476  & \underline{0.450}\\
    
    \midrule
    \multicolumn{16}{c}{\textbf{DSTAGNN-NB Outputs}} \\
    \midrule
    CCR\_1h     & 0.328 &     0.695  &    0.695 &       0.958 &  0.667 & \underline{0.230} &    \textbf{0.107} &  \vline  & 0.292 &     0.773 &    0.752 &       4.853 &  0.900 &  \underline{0.120}  & \textbf{0.102}\\
    CCR\_8h      &    0.895 &     0.941 &    0.940 &       0.604 &  0.928 &  \underline{0.443}  &  \textbf{0.440}  & \vline  &         0.811 &     0.905 &    0.901 &       \underline{0.419} &  0.930 & 0.572  &  \textbf{0.314} \\
    CCR\_1d       &  0.962 &     0.950 &    0.950 &       0.810 &  0.946 & \underline{0.792}  &   \textbf{0.709} &  \vline  &  0.822 &     0.790 &    0.788 &       0.619 &  0.797 &  \underline{0.235}  & \textbf{0.171} \\
    CCR\_1w        &   0.999 &     1.000 &    1.000 &       0.987 &  1.000 &  \underline{0.977} &   \textbf{0.952}   &  \vline  &  0.260  &     0.140 &    0.140 &       \textbf{0.138} &  \underline{0.139} & 0.150 &  0.149 \\
    
    \midrule
    CTC\_1h     &    0.398 &     0.187 &    \underline{0.121} &     0.138 &  0.129 & 0.159 & \textbf{0.119} &  \vline  & 0.937 &     0.652 &    0.661 &     1.222 &  0.970 & \underline{0.631} & \textbf{0.229} \\
    CTC\_8h       &   1.462 &     0.666 &    0.669 &      \underline{0.473} &  0.767 & 0.505 & \textbf{0.452} & \vline  & 0.694 &     \underline{0.575} &    0.608 &       0.586 &  0.950 & 0.612 & \textbf{0.523} \\
    CTC\_1d     &   1.393 &     0.776 &    0.774 &       1.304 &  0.725 & \underline{0.481} &  \textbf{0.402} &  \vline  & 1.650 &     0.789 &    0.771 &       1.259 &  1.000 & \underline{0.637} & \textbf{0.503}  \\
    CTC\_1w       & 0.890 &     0.876 &    0.874 &       \textbf{0.717} &  0.840 &   \underline{0.824} & 0.833  & \vline  &  0.980 &     0.757 &    0.743 &       \textbf{0.498} &  0.827 & 0.679 &  \underline{0.550}\\
    \bottomrule
    \end{tabularx}
    \caption{ENCE of the calibration methods. Bold fonts mark the best and underlines denote the second-best calibration results.}
    \label{tab:ce_full}
\end{table*}

\subsection{Baseline Calibration Methods}

We compare the SAUC framework with existing post-hoc calibration methods proven to be effective in the regression tasks:
(1) \textit{Isotonic regression} \citep{niculescu2005predicting}, which ensures non-decreasing predictive probabilities, assuming higher model accuracy with increased predicted probabilities; 
(2) \textit{Temperature scaling} \citep{kull2019beyond}, which scales the model's output using a learned parameter, whose efficacy for regression models depends on the output's monotonicity in relation to confidence; 
(3) \textit{Histogram binning} \citep{zadrozny2001obtaining}, a model partitioning predicted values into bins, calibrating each independently according to observed frequencies, yet might lack flexibility for diverse prediction intervals; 
(4) \textit{Platt scaling} \citep{kuleshov2018accurate}, which applies a scaling transformation to the predicted values with the function derived from minimizing observed and predicted value discrepancies, providing linear adaptability but potentially inadequate for complex uncertainty patterns; 
(5) \textit{Quantile regression} \citep{chung2021beyond}, see Section \ref{sec:qr}.

\subsection{Calibration Results Comparison}


\subsubsection{ENCE Evaluation}
Table \ref{tab:ce_full} presents the calibration performance of baseline models and our proposed framework in different cases, measured by ENCE. 
The term "zero-only targets" denotes ENCE computation solely on true zero target values.

Table \ref{tab:ce_full} highlights the superior performance of the SAUC framework across most datasets, with bold type indicating the best results and underlines denoting the second-best.
Our findings reveal that the SAUC framework typically outperforms other models, though it might fall short in aggregated cases, such as those with a 1-week resolution. 
Notably, SAUC demonstrates approximately an overall 23\% reduction in ENCE compared to the second-best model with full observations and a similar 20\% reduction in sparsity entries. 
This reduction is assessed by computing the average percentage improvement relative to the second-best model across all scenarios.
The separation of zero and non-zero data during calibration proved particularly beneficial, effectively addressing the issues of heteroscedasticity and zero inflation.
However, some calibration methods may worsen errors due to overfitting, particularly evident in temperature scaling and isotonic regression.

Histogram binning and isotonic regression exhibit similar results among baseline models, facilitated by the non-decreasing nature of Equation \ref{eq:cdf}. 
Temperature scaling performs well in coarse temporal resolutions, while Platt scaling excels in sparse scenarios. QR consistently ranks second-best, with our SAUC framework closely matching QR in aggregated non-sparse cases and outperforming notably in sparse scenarios.

\subsubsection{Reliability Diagram Evaluation}
\begin{figure*}[!ht]
    \centering    
    \subfigure[Full observations, CCR\_8h]{\includegraphics[scale=1]{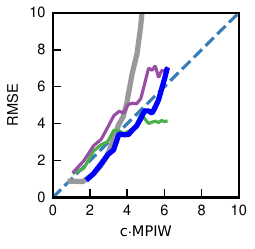}\label{fig:ccr8h_full}}
    \subfigure[Zero-only targets, CCR\_8h]{\includegraphics[scale=1]{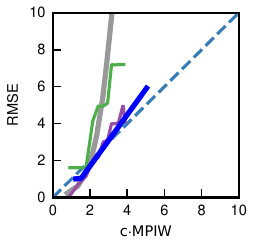}\label{fig:ccr8h_zero}}
    \subfigure[Full observations, CCR\_1h]{\includegraphics[scale=1]{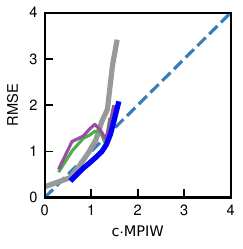}\label{fig:ccr1h_full}}
    \subfigure[Zero-only targets, CCR\_1h]{\includegraphics[scale=1]{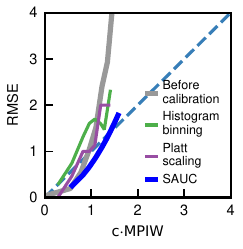}\label{fig:ccr1h_zero}}    
    \caption{Reliability diagrams of different calibration methods applied to STGCN-NB outputs on different components of CCR\_8h and CCR\_1h data. Results closer to the diagonal dashed line are considered better.}
    \label{fig:ccr8h}
\end{figure*}
Utilizing Equations~\ref{eq:define_calib} and~\ref{eq:ence}, we refine the conventional reliability diagram to evaluate the calibration efficacy of a model \citep{kull2019beyond}. 
This involves partitioning predictions into bins based on $\mathcal{I}$, after which we compute the RMSE and MPIW employing the calibrated $\mu^*$ and $|\mathcal{I}^*|$ for each respective bin.

Figure \ref{fig:ccr8h} presents a comparative reliability diagram of STGCN-NB outputs prior to calibration, alongside Isotonic regression, Platt scaling, and our SAUC framework, each applied to different segments of the CCR\_8h dataset. 
We selectively highlight representative non-parametric and parametric baseline calibration approaches based on their performance. 
The diagonal dashed line symbolizes the calibration ideal: a closer alignment to this line indicates superior calibration.


A closer examination of Figure \ref{fig:ccr8h_full} reveals that both baseline models and our SAUC framework adeptly calibrate the model outputs across all observations. 
Yet, focusing solely on the reliability diagram of our calibrated outputs for ground-truth zero values, as depicted in Figure \ref{fig:ccr8h_zero}, there is a noticeable deviation from the ideal line. 
Conversely, our SAUC framework demonstrates commendable alignment with the ideal when authentic data are zeros. 
This observation remains consistent for the CCR\_1h dataset, as seen in Figures \ref{fig:ccr1h_full} and \ref{fig:ccr1h_zero}. 
By contrasting Figures \ref{fig:ccr8h_full} and \ref{fig:ccr1h_full}, it becomes evident that the calibration performance diminishes in scenarios with finer resolutions. 
Nevertheless, our SAUC framework retains commendable efficacy, particularly when predicting zero entries, attributed to our distinctive calibration approach coupled with QR's adeptness at managing heteroscedasticity, especially when predictions on zeros exhibit considerable variability. 
Contemporary probabilistic spatiotemporal prediction models predominantly prioritize refining MPIWs for enhanced accuracy, often unintentionally neglecting practical safety considerations. 
Our calibration strategy offers outputs with reinforced reliability.

\subsubsection{Coverage Evaluation}
\begin{figure}[htbp]
    \centering
    \subfigure[Calibration on CCR dataset]{\includegraphics[width=0.42\textwidth]{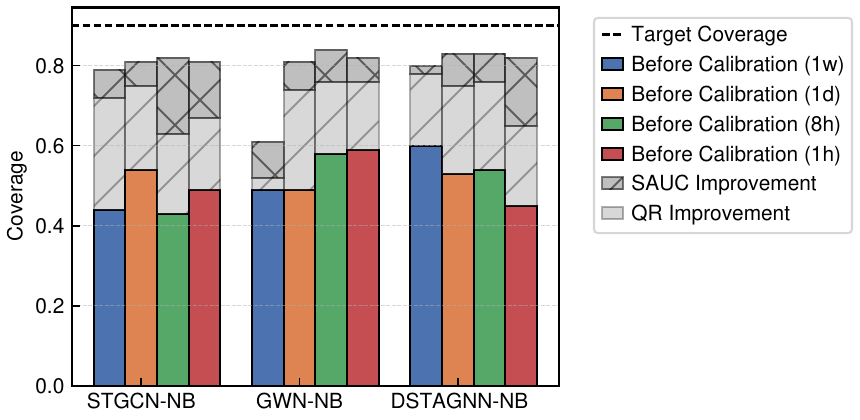}\label{fig:coverage_crime}} \hfill 
    \subfigure[Calibration on CTC dataset]{\includegraphics[width=0.42\textwidth]{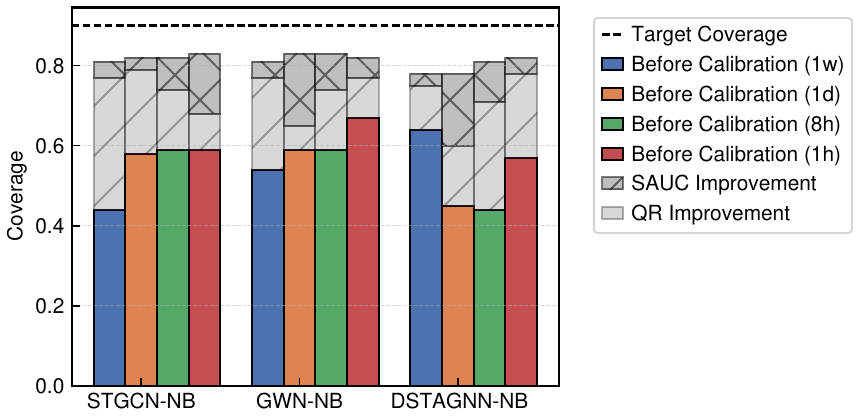}\label{fig:coverage_crash}} 
    \caption{Coverage improvements by applying SAUC framework and QR on different data sets and models.}
    \label{fig:coverage}
\end{figure}

We further evaluate the calibration performance through coverage, defined in Equation \ref{eq:coverage}. 
The calibration improvements achieved using the SAUC framework and QR are illustrated in Figure \ref{fig:coverage}, applied to various pre-calibrated models across different resolutions of the CCR and CTC datasets. 
The QR improvement in Figure \ref{fig:coverage} signifies the enhancement in coverage by applying QR to the pre-calibrated model, while SAUC improvement denotes the further enhancement over QR. 
Across all models and resolutions, calibration markedly enhances coverage. 
Initially, coverage levels were consistently below the target (0.9), exhibiting notable variations depending on the temporal resolution. 
Following calibration, all models and datasets achieve a similar level of coverage, closer to the target. 
Particularly on the sparser datasets (resolutions 1h and 8h), the improvements of SAUC compared with QR are generally more pronounced, notably for model performance on the CCR datasets. 
These findings underscore the efficacy of the SAUC framework in enhancing model calibration and ensuring improved prediction reliability across diverse temporal resolutions and models, particularly when handling sparser data inputs.

\subsection{Risk Analysis}
To evaluate the spatial effects of spatiotemporal prediction after SAUC calibration, we introduce Risk Score (RS), an enhancement of the traditional metric defined as the product of \textit{event probability} and \textit{potential loss magnitude} \citep{ng2002algorithm,vose2008risk}: 
\begin{equation}
    RS = \hat{\mu} \times |\hat{\mathcal{I}}|.   
\end{equation}
This metric integrates anticipated risk ($\hat{\mu}$) and prediction uncertainty ($|\hat{\mathcal{I}}|$), attributing higher RS to regions with both high incident frequency and uncertainty. 
This RS metric assigns higher risk to predictions that are characterized by both large predicted mean values and high uncertainty.
\begin{figure}[!ht]
    \centering
    \subfigure[Avg. num. of crashes]{\includegraphics[width=0.23\textwidth]{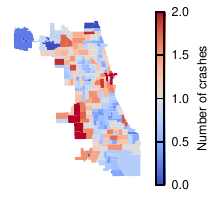}\label{fig:case_study_expect}}   \hfill 
    \subfigure[RS distribution using prediction before calibration]{\includegraphics[width=0.23\textwidth]{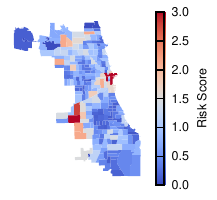}\label{fig:case_study_before}} \hfill
    \subfigure[RS distribution using calibrated prediction via Platt scaling]{\includegraphics[width=0.23\textwidth]{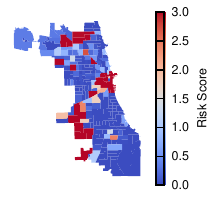}\label{fig:case_study_baseline}} \hfill
    \subfigure[RS distribution using using calibrated prediction via SAUC]{\includegraphics[width=0.23\textwidth]{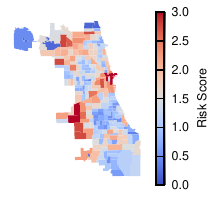}\label{fig:case_study_calib}}
    \caption{Traffic crash accident distributions and calibration using CTC\_8h data. All the data are averaged over the temporal dimension.}
    \label{fig:case_study}
\end{figure}

Using the CTC\_8h dataset and outputs from STGCN-NB, we demonstrate the importance of SAUC for evaluating the risks of traffic accidents. 
Figure \ref{fig:case_study} compares the average number of crashes and average RS values over time for each region, pre and post-calibration. 
Prior to calibration, as depicted in Figure \ref{fig:case_study_before}, the RS in regions, especially Chicago's northern and southern sectors, did not provide actionable insights, given their frequent traffic incidents reported by Figure \ref{fig:case_study_expect}. 
This is due to the tight PIs from pre-calibration method outputs lower the RS. 
As shown in Figure \ref{fig:case_study_baseline}, RS distribution from Platt scaling calibrated predictions could highlight certain risky areas, but still miss the major crash regions in southern Chicago. 
In contrast, SAUC-calibrated predictions, as seen in Figure \ref{fig:case_study_calib}, unveil coherent spatial distribution patterns highly consistent with Figure \ref{fig:case_study_expect}, demonstrating the effectiveness for urban safety monitoring.

Rather than just yielding exact numerical predictions with tight confidence intervals, it is imperative that models convey equivalent severity levels when integrating uncertainty. 
Such an approach ensures reliable outputs in crime and accident predictions. 
Notably, the reliable estimation of zero or non-zero incident counts is pivotal in risk management. 
Existing probabilistic prediction studies focus on exact values and restricted confidence intervals, sometimes neglecting the data variability, especially around zero incidences.

\subsection{Sensitivity Analysis}
\label{sec:num_bins}
The number of bins $N$ in both the SAUC framework and ENCE discussion is important.
We choose the bin number from the range $N$ = [1, 5, 10, 15, 20, 25, 30] and discuss the ENCE change with respect to the bin number. 
We conduct the sensitivity analysis using the four cases in the crime dataset using STGCN-NB model outputs.

\begin{table}[h!]
\centering
\begin{tabular}{ccccccc}
\hline
\textbf{Dataset/Bin} & \textbf{1} & \textbf{5} & \textbf{10} & \textbf{15} & \textbf{20} & \textbf{25} \\
\hline
\multicolumn{7}{c}{\textbf{Full Observations}} \\
\hline
CCR\_1h & 0.513 & 0.302 & 0.299 & 0.198 & 0.249 & 0.292 \\
CCR\_8h & 0.096 & 0.113 & 0.185 & 0.336 & 0.123 & 0.115 \\
CCR\_1d & 0.093 & 0.167 & 0.185 & 0.192 & 0.189 & 0.207 \\
CCR\_1w & 0.074 & 0.216 & 0.323 & 0.242 & 0.254 & 0.272 \\
\hline
\multicolumn{7}{c}{\textbf{Zero-only Observations}} \\
\hline
CCR\_1h & 0.650 & 0.689 & 0.604 & 0.267 & 0.229 & 0.501 \\
CCR\_8h & 0.642 & 0.583 & 0.504 & 0.127 & 0.172 & 0.345 \\
CCR\_1d & 0.641 & 1.181 & 1.096 & 0.738 & 1.025 & 0.958 \\
CCR\_1w & 1.445 & 0.950 & 0.361 & 0.113 & 0.363 & 0.567 \\
\hline
\end{tabular}
\caption{ENCE on the full observations and zero-only observations with respect to different bin numbers.}
\label{tab:combined_sensitivity_bins}
\end{table}

From Table \ref{tab:combined_sensitivity_bins}, it is evident that the number of bins significantly impacts the SAUC framework and ENCE performance. 
Both very small and very large bin numbers may result in suboptimal ENCE values. 
Intuitively, with only one bin, all data points are grouped together, making it difficult to differentiate levels of variance as the comparison involves the overall data variability and the average MPIW. 
Conversely, having too many bins means each bin contains few data points, leading to fluctuations due to the small sample size. 
Therefore, an optimal bin number range between 10 and 20 is preferable for achieving reliable ENCE.

\subsection{Discussion on Distribution Assumptions}
\label{sec:poisson}
A common implementation question arises: is the NB distribution flexible enough to capture the response distributions? 
The focus of this work lies in uncertainty calibration methods for sparse data, where specific selections of distributions are less discussed.
Nevertheless, it is imperative to examine whether the presented distributions offer sufficient flexibility. 

In comparison to the NB distribution, we have also implemented Poisson and Gaussian distributions as estimated distributions for our model. 
Models' modification remains unchanged following Section \ref{sec:stgnn}, with Equation \ref{eq:PGNN} replaced by estimations of corresponding parameters for Poisson and Gaussian distributions. 
The last layer is modified to output the distribution parameters of the Poisson distributions exclusively. 
For the Poisson distribution, only one parameter $\lambda$ is outputted, and the loss function can be similarly defined as:

\begin{equation}
\begin{aligned}
\mathcal{L}(\hat{\lambda}, y) &= \frac{1}{N} \sum_{i=1}^N \left( \hat{\lambda}_{s,i} - y_i \cdot \log(\lambda_{s,i}) \right),
\label{eq:poisson_loss}
\end{aligned}
\end{equation}
where $\hat{\lambda}_{s} = \hat{\lambda} + \epsilon$, with \(\hat{\lambda}\) representing the predicted rate parameters, \(\text{target}\) representing the target counts, and \(\epsilon\) a small constant (e.g., \(10^{-10}\)) added for numerical stability. 
Equation \ref{eq:poisson_loss} is derived from the Poisson likelihood function, ensuring robust and stable learning for our spatiotemporal prediction models. 

Similarly, for Gaussian distribution, we estimate the location and scale parameters $\mu$ and $\sigma$ with the loss function:
\begin{equation}
\begin{aligned}
\mathcal{L}(\hat{\mu}, \hat{\sigma}, y) &= \frac{1}{N} \sum_{i=1}^N \left( \frac{1}{2} \log(2\pi \hat{\sigma}_i^2) +  \frac{(y_i - \hat{\mu}_i)^2}{2\hat{\sigma}_i^2} \right) + \lambda||\hat{\sigma}||_2 ,
\label{eq:gaussian_loss}
\end{aligned}
\end{equation}
where penalization for the variance is also added, similar to Equation \ref{eq:modify_loss}. 
We discuss the prediction RMSE and ENCE performance of STGCN, STGCN-NB, STGCN-Poisson, and STGCN-Gaussian, as shown in Table \ref{tab:poisson_vs_nb}.
\begin{table}[h!]
\centering
\resizebox{\columnwidth}{!}{
\begin{tabular}{ccccccc}
\hline
\textbf{Dataset} & \textbf{Metric} & \textbf{Original} & \textbf{NB} & \textbf{Poisson} & \textbf{Gaussian} \\
\hline
\multirow{2}{*}{CCR\_1h} & RMSE & 0.637 & 0.703 & 0.581 & 0.741 \\
 & ENCE & / & 1.160 & 1.692 & 3.292 \\
\hline
\multirow{2}{*}{CCR\_8h} & RMSE & 2.386 & 2.326 & 2.419 & 2.303 \\
 & ENCE & / & 1.389 & 0.913 & 30.286 \\
\hline
\multirow{2}{*}{CCR\_1d} & RMSE & 5.167 & 4.286 & 6.296 & 6.701 \\
 & ENCE & / & 0.413 & 0.999 & 67.267 \\
\hline
\multirow{2}{*}{CCR\_1w} & RMSE & 23.333 & 20.242 & 23.907 & 31.044 \\
 & ENCE & / & 0.855 & 0.100 & 168.889 \\
\hline
\multirow{2}{*}{CTC\_1h} & RMSE & 0.214 & 0.220 & 0.282 & 0.321 \\
 & ENCE & / & 0.578 & 3.079 & 0.962 \\
\hline
\multirow{2}{*}{CTC\_8h} & RMSE & 0.633 & 0.726 & 0.962 & 0.972 \\
 & ENCE & / & 0.323 & 0.960 & 2.320 \\
\hline
\multirow{2}{*}{CTC\_1d} & RMSE & 1.148 & 1.786 & 1.138 & 1.130 \\
 & ENCE & / & 0.722 & 1.509 & 3.351 \\
\hline
\multirow{2}{*}{CTC\_1w} & RMSE & 3.317 & 3.929 & 3.450 & 3.321 \\
 & ENCE & / & 2.279 & 0.145 & 5.867 \\
\hline
\end{tabular}}
\caption{Comparison of RMSE and ENCE for the original model and its NB, Poisson, and Gaussian modification.}
\label{tab:poisson_vs_nb}
\end{table}

Based on Table \ref{tab:poisson_vs_nb}, it is clear that the Poisson and Gaussian modifications exhibit similar performance to the NB distribution in terms of RMSE based on the experimental results. 
However, NB demonstrates better ENCE when capturing sparse data cases compared with the other two distributions, thus it is selected as the distribution predominantly discussed in this paper. 
More dedicated data distributions' assumptions can be explored, as demonstrated by the work of \citep{jiang2023uncertainty,zhuang2022uncertainty,zuo2020transformer,bazzani2016recurrent}.

\subsection{Complexity Discussion}
\begin{figure}[ht]
    \centering    \includegraphics[width=0.8\linewidth]{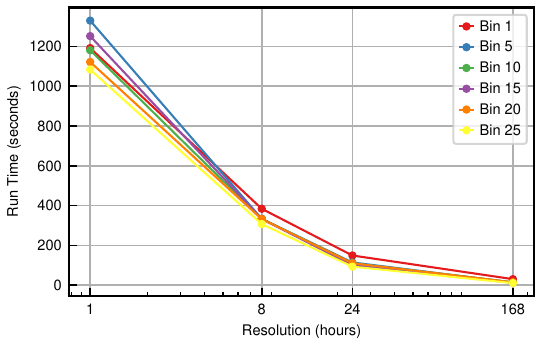}
    \caption{Runtime of the calibration methods on the full observations with different bin numbers and data resolutions of CCR dataset. The x-axis scale is logarithmic.}   
    \label{fig:runtime}
\end{figure}

From Equation \ref{eq:QR}, we know that the complexity of the SAUC framework depends on the size of the data.
Since our method is a post-hoc method, the added complexity is easier to address in terms of runtime.
Same as the settings in sensitivity analysis, we visualize the model runtime on different bin numbers and data resolutions by running STGCN-NB, shown in Figure \ref{fig:runtime}.
It is clear that the selection of the bin number does not influence the complexity and runtime.
Moreover, by reducing the resolution from 1 hour to 8 hours, 1 day, and 1 week, the runtime decreases nearly exponentially with the change in the temporal resolutions of the data.
\section{Conclusion}
Current spatiotemporal GNN models mainly yield deterministic predictions, overlooking data uncertainties.
Given the sparsity and asymmetry in high-resolution spatiotemporal data, quantifying uncertainty is challenging. 
Previous ST-GNN models covert the deterministic models to probabilistic ones by changing the output layers, without proper discussions in the reasonableness of learned distributions and UQ.
We introduce the SAUC framework, which calibrates uncertainties for both zero and non-zero values using the zero-inflated nature of these datasets.
As a post-hoc method, SAUC is available to be adapted to any existing ST-GNN model.
We further introduce new calibration metrics suited for asymmetric distributions. 
Tests on real-world datasets show SAUC reduces calibration errors by 20\% for zero-only targets. 
SAUC enhances GNN models and aids risk assessment in sparse datasets where accurate predictions are critical due to safety concerns.

\section*{Acknowledgments}
We extend our gratitude to the reviewers for their time and effort in reviewing and refining our manuscript.

This material is based upon work supported by the U.S. Department of Energy’s Office of Energy Efficiency and Renewable Energy (EERE) under the Vehicle Technology Program Award Number DE-EE0009211.

Shenhao Wang and Yuheng Bu acknowledge the support from the Research Opportunity Seed Fund (ROSF) 2023 at the University of Florida.
Guang Wang is partially supported by the National Science Foundation under Grant No. 2411152

\clearpage
\bibliographystyle{ACM-Reference-Format}
\bibliography{acmart}


\begin{thebibliography}{48}


\ifx \showCODEN    \undefined \def \showCODEN     #1{\unskip}     \fi
\ifx \showDOI      \undefined \def \showDOI       #1{#1}\fi
\ifx \showISBNx    \undefined \def \showISBNx     #1{\unskip}     \fi
\ifx \showISBNxiii \undefined \def \showISBNxiii  #1{\unskip}     \fi
\ifx \showISSN     \undefined \def \showISSN      #1{\unskip}     \fi
\ifx \showLCCN     \undefined \def \showLCCN      #1{\unskip}     \fi
\ifx \shownote     \undefined \def \shownote      #1{#1}          \fi
\ifx \showarticletitle \undefined \def \showarticletitle #1{#1}   \fi
\ifx \showURL      \undefined \def \showURL       {\relax}        \fi
\providecommand\bibfield[2]{#2}
\providecommand\bibinfo[2]{#2}
\providecommand\natexlab[1]{#1}
\providecommand\showeprint[2][]{arXiv:#2}

\bibitem[Abdo et~al\mbox{.}(2017)]%
        {abdo2017uncertainty}
\bibfield{author}{\bibinfo{person}{H Abdo}, \bibinfo{person}{Jean-Marie Flaus}, {and} \bibinfo{person}{F Masse}.} \bibinfo{year}{2017}\natexlab{}.
\newblock \showarticletitle{Uncertainty quantification in risk assessment-representation, propagation and treatment approaches: application to atmospheric dispersion modeling}.
\newblock \bibinfo{journal}{\emph{Journal of Loss Prevention in the Process Industries}}  \bibinfo{volume}{49} (\bibinfo{year}{2017}), \bibinfo{pages}{551--571}.
\newblock


\bibitem[Agarwal et~al\mbox{.}(2002)]%
        {agarwal2002zero}
\bibfield{author}{\bibinfo{person}{Deepak~K Agarwal}, \bibinfo{person}{Alan~E Gelfand}, {and} \bibinfo{person}{Steven Citron-Pousty}.} \bibinfo{year}{2002}\natexlab{}.
\newblock \showarticletitle{Zero-inflated models with application to spatial count data}.
\newblock \bibinfo{journal}{\emph{Environmental and Ecological statistics}}  \bibinfo{volume}{9} (\bibinfo{year}{2002}), \bibinfo{pages}{341--355}.
\newblock


\bibitem[Antunes et~al\mbox{.}(2017)]%
        {antunes2017review}
\bibfield{author}{\bibinfo{person}{Francisco Antunes}, \bibinfo{person}{Aidan O’Sullivan}, \bibinfo{person}{Filipe Rodrigues}, {and} \bibinfo{person}{Francisco Pereira}.} \bibinfo{year}{2017}\natexlab{}.
\newblock \showarticletitle{A review of heteroscedasticity treatment with Gaussian processes and quantile regression meta-models}.
\newblock \bibinfo{journal}{\emph{Seeing Cities Through Big Data: Research, Methods and Applications in Urban Informatics}} (\bibinfo{year}{2017}), \bibinfo{pages}{141--160}.
\newblock


\bibitem[Bazzani et~al\mbox{.}(2016)]%
        {bazzani2016recurrent}
\bibfield{author}{\bibinfo{person}{Loris Bazzani}, \bibinfo{person}{Hugo Larochelle}, {and} \bibinfo{person}{Lorenzo Torresani}.} \bibinfo{year}{2016}\natexlab{}.
\newblock \showarticletitle{Recurrent mixture density network for spatiotemporal visual attention}.
\newblock \bibinfo{journal}{\emph{arXiv preprint arXiv:1603.08199}} (\bibinfo{year}{2016}).
\newblock


\bibitem[Bi et~al\mbox{.}(2023)]%
        {bi2023mm}
\bibfield{author}{\bibinfo{person}{Wendong Bi}, \bibinfo{person}{Lun Du}, \bibinfo{person}{Qiang Fu}, \bibinfo{person}{Yanlin Wang}, \bibinfo{person}{Shi Han}, {and} \bibinfo{person}{Dongmei Zhang}.} \bibinfo{year}{2023}\natexlab{}.
\newblock \showarticletitle{MM-GNN: Mix-Moment Graph Neural Network towards Modeling Neighborhood Feature Distribution}. In \bibinfo{booktitle}{\emph{Proceedings of the Sixteenth ACM International Conference on Web Search and Data Mining}}. \bibinfo{pages}{132--140}.
\newblock


\bibitem[Bishop(2006)]%
        {bishop2006pattern}
\bibfield{author}{\bibinfo{person}{C Bishop}.} \bibinfo{year}{2006}\natexlab{}.
\newblock \showarticletitle{Pattern recognition and machine learning}.
\newblock \bibinfo{journal}{\emph{Springer google schola}}  \bibinfo{volume}{2} (\bibinfo{year}{2006}), \bibinfo{pages}{531--537}.
\newblock


\bibitem[Chen and Tokdar(2021)]%
        {chen2021joint}
\bibfield{author}{\bibinfo{person}{Xu Chen} {and} \bibinfo{person}{Surya~T Tokdar}.} \bibinfo{year}{2021}\natexlab{}.
\newblock \showarticletitle{Joint quantile regression for spatial data}.
\newblock \bibinfo{journal}{\emph{Journal of the Royal Statistical Society Series B: Statistical Methodology}} \bibinfo{volume}{83}, \bibinfo{number}{4} (\bibinfo{year}{2021}), \bibinfo{pages}{826--852}.
\newblock


\bibitem[Chung et~al\mbox{.}(2021)]%
        {chung2021beyond}
\bibfield{author}{\bibinfo{person}{Youngseog Chung}, \bibinfo{person}{Willie Neiswanger}, \bibinfo{person}{Ian Char}, {and} \bibinfo{person}{Jeff Schneider}.} \bibinfo{year}{2021}\natexlab{}.
\newblock \showarticletitle{Beyond pinball loss: Quantile methods for calibrated uncertainty quantification}.
\newblock \bibinfo{journal}{\emph{Advances in Neural Information Processing Systems}}  \bibinfo{volume}{34} (\bibinfo{year}{2021}), \bibinfo{pages}{10971--10984}.
\newblock


\bibitem[Fernandes et~al\mbox{.}(2009)]%
        {fernandes2009modelling}
\bibfield{author}{\bibinfo{person}{Marcus~VM Fernandes}, \bibinfo{person}{Alexandra~M Schmidt}, {and} \bibinfo{person}{Helio~S Migon}.} \bibinfo{year}{2009}\natexlab{}.
\newblock \showarticletitle{Modelling zero-inflated spatio-temporal processes}.
\newblock \bibinfo{journal}{\emph{Statistical Modelling}} \bibinfo{volume}{9}, \bibinfo{number}{1} (\bibinfo{year}{2009}), \bibinfo{pages}{3--25}.
\newblock


\bibitem[Gao et~al\mbox{.}(2023)]%
        {gao2023spatiotemporal}
\bibfield{author}{\bibinfo{person}{Xiaowei Gao}, \bibinfo{person}{Huanfa Chen}, {and} \bibinfo{person}{James Haworth}.} \bibinfo{year}{2023}\natexlab{}.
\newblock \showarticletitle{A spatiotemporal analysis of the impact of lockdown and coronavirus on London’s bicycle hire scheme: from response to recovery to a new normal}.
\newblock \bibinfo{journal}{\emph{Geo-spatial Information Science}} (\bibinfo{year}{2023}), \bibinfo{pages}{1--21}.
\newblock


\bibitem[Gawlikowski et~al\mbox{.}(2021)]%
        {gawlikowski2021survey}
\bibfield{author}{\bibinfo{person}{Jakob Gawlikowski}, \bibinfo{person}{Cedrique Rovile~Njieutcheu Tassi}, \bibinfo{person}{Mohsin Ali}, \bibinfo{person}{Jongseok Lee}, \bibinfo{person}{Matthias Humt}, \bibinfo{person}{Jianxiang Feng}, \bibinfo{person}{Anna Kruspe}, \bibinfo{person}{Rudolph Triebel}, \bibinfo{person}{Peter Jung}, \bibinfo{person}{Ribana Roscher}, {et~al\mbox{.}}} \bibinfo{year}{2021}\natexlab{}.
\newblock \showarticletitle{A survey of uncertainty in deep neural networks}.
\newblock \bibinfo{journal}{\emph{arXiv preprint arXiv:2107.03342}} (\bibinfo{year}{2021}).
\newblock


\bibitem[Gruber and Buettner(2022)]%
        {gruber2022better}
\bibfield{author}{\bibinfo{person}{Sebastian Gruber} {and} \bibinfo{person}{Florian Buettner}.} \bibinfo{year}{2022}\natexlab{}.
\newblock \showarticletitle{Better uncertainty calibration via proper scores for classification and beyond}.
\newblock \bibinfo{journal}{\emph{Advances in Neural Information Processing Systems}}  \bibinfo{volume}{35} (\bibinfo{year}{2022}), \bibinfo{pages}{8618--8632}.
\newblock


\bibitem[Hao and Naiman(2007)]%
        {hao2007quantile}
\bibfield{author}{\bibinfo{person}{Lingxin Hao} {and} \bibinfo{person}{Daniel~Q Naiman}.} \bibinfo{year}{2007}\natexlab{}.
\newblock \bibinfo{booktitle}{\emph{Quantile regression}}.
\newblock Number 149. \bibinfo{publisher}{Sage}.
\newblock


\bibitem[Huang et~al\mbox{.}(2023)]%
        {huang_uncertainty_2023}
\bibfield{author}{\bibinfo{person}{Kexin Huang}, \bibinfo{person}{Ying Jin}, \bibinfo{person}{Emmanuel Candes}, {and} \bibinfo{person}{Jure Leskovec}.} \bibinfo{year}{2023}\natexlab{}.
\newblock \showarticletitle{Uncertainty quantification over graph with conformalized graph neural networks}.
\newblock \bibinfo{journal}{\emph{NeurIPS}} (\bibinfo{year}{2023}).
\newblock


\bibitem[Jiang et~al\mbox{.}(2023)]%
        {jiang2023uncertainty}
\bibfield{author}{\bibinfo{person}{Xinke Jiang}, \bibinfo{person}{Dingyi Zhuang}, \bibinfo{person}{Xianghui Zhang}, \bibinfo{person}{Hao Chen}, \bibinfo{person}{Jiayuan Luo}, {and} \bibinfo{person}{Xiaowei Gao}.} \bibinfo{year}{2023}\natexlab{}.
\newblock \showarticletitle{Uncertainty Quantification via Spatial-Temporal Tweedie Model for Zero-inflated and Long-tail Travel Demand Prediction}. In \bibinfo{booktitle}{\emph{Proceedings of the 32nd ACM International Conference on Information and Knowledge Management}} (, Birmingham, United Kingdom,) \emph{(\bibinfo{series}{CIKM '23})}. \bibinfo{publisher}{Association for Computing Machinery}, \bibinfo{address}{New York, NY, USA}, \bibinfo{pages}{3983–3987}.
\newblock
\showISBNx{9798400701245}
\urldef\tempurl%
\url{https://doi.org/10.1145/3583780.3615215}
\showDOI{\tempurl}


\bibitem[Khosravi et~al\mbox{.}(2010)]%
        {khosravi2010lower}
\bibfield{author}{\bibinfo{person}{Abbas Khosravi}, \bibinfo{person}{Saeid Nahavandi}, \bibinfo{person}{Doug Creighton}, {and} \bibinfo{person}{Amir~F Atiya}.} \bibinfo{year}{2010}\natexlab{}.
\newblock \showarticletitle{Lower upper bound estimation method for construction of neural network-based prediction intervals}.
\newblock \bibinfo{journal}{\emph{IEEE transactions on neural networks}} \bibinfo{volume}{22}, \bibinfo{number}{3} (\bibinfo{year}{2010}), \bibinfo{pages}{337--346}.
\newblock


\bibitem[Kuleshov et~al\mbox{.}(2018)]%
        {kuleshov2018accurate}
\bibfield{author}{\bibinfo{person}{Volodymyr Kuleshov}, \bibinfo{person}{Nathan Fenner}, {and} \bibinfo{person}{Stefano Ermon}.} \bibinfo{year}{2018}\natexlab{}.
\newblock \showarticletitle{Accurate uncertainties for deep learning using calibrated regression}. In \bibinfo{booktitle}{\emph{International conference on machine learning}}. PMLR, \bibinfo{pages}{2796--2804}.
\newblock


\bibitem[Kull et~al\mbox{.}(2019)]%
        {kull2019beyond}
\bibfield{author}{\bibinfo{person}{Meelis Kull}, \bibinfo{person}{Miquel Perello~Nieto}, \bibinfo{person}{Markus K{\"a}ngsepp}, \bibinfo{person}{Telmo Silva~Filho}, \bibinfo{person}{Hao Song}, {and} \bibinfo{person}{Peter Flach}.} \bibinfo{year}{2019}\natexlab{}.
\newblock \showarticletitle{Beyond temperature scaling: Obtaining well-calibrated multi-class probabilities with dirichlet calibration}.
\newblock \bibinfo{journal}{\emph{Advances in neural information processing systems}}  \bibinfo{volume}{32} (\bibinfo{year}{2019}).
\newblock


\bibitem[Kumar et~al\mbox{.}(2019)]%
        {kumar2019verified}
\bibfield{author}{\bibinfo{person}{Ananya Kumar}, \bibinfo{person}{Percy~S Liang}, {and} \bibinfo{person}{Tengyu Ma}.} \bibinfo{year}{2019}\natexlab{}.
\newblock \showarticletitle{Verified uncertainty calibration}.
\newblock \bibinfo{journal}{\emph{Advances in Neural Information Processing Systems}}  \bibinfo{volume}{32} (\bibinfo{year}{2019}).
\newblock


\bibitem[Lan et~al\mbox{.}(2022)]%
        {lan2022dstagnn}
\bibfield{author}{\bibinfo{person}{Shiyong Lan}, \bibinfo{person}{Yitong Ma}, \bibinfo{person}{Weikang Huang}, \bibinfo{person}{Wenwu Wang}, \bibinfo{person}{Hongyu Yang}, {and} \bibinfo{person}{Pyang Li}.} \bibinfo{year}{2022}\natexlab{}.
\newblock \showarticletitle{Dstagnn: Dynamic spatial-temporal aware graph neural network for traffic flow forecasting}. In \bibinfo{booktitle}{\emph{International conference on machine learning}}. PMLR, \bibinfo{pages}{11906--11917}.
\newblock


\bibitem[Laves et~al\mbox{.}(2020)]%
        {laves2020well}
\bibfield{author}{\bibinfo{person}{Max-Heinrich Laves}, \bibinfo{person}{Sontje Ihler}, \bibinfo{person}{Jacob~F Fast}, \bibinfo{person}{L{\"u}der~A Kahrs}, {and} \bibinfo{person}{Tobias Ortmaier}.} \bibinfo{year}{2020}\natexlab{}.
\newblock \showarticletitle{Well-calibrated regression uncertainty in medical imaging with deep learning}. In \bibinfo{booktitle}{\emph{Medical Imaging with Deep Learning}}. PMLR, \bibinfo{pages}{393--412}.
\newblock


\bibitem[Levi et~al\mbox{.}(2020)]%
        {levi_evaluating_2020}
\bibfield{author}{\bibinfo{person}{Dan Levi}, \bibinfo{person}{Liran Gispan}, \bibinfo{person}{Niv Giladi}, {and} \bibinfo{person}{Ethan Fetaya}.} \bibinfo{year}{2020}\natexlab{}.
\newblock \bibinfo{title}{Evaluating and {Calibrating} {Uncertainty} {Prediction} in {Regression} {Tasks}}.
\newblock
\newblock
\urldef\tempurl%
\url{http://arxiv.org/abs/1905.11659}
\showURL{%
\tempurl}
\newblock
\shownote{arXiv:1905.11659 [cs, stat]}.


\bibitem[Levi et~al\mbox{.}(2022)]%
        {levi2022evaluating}
\bibfield{author}{\bibinfo{person}{Dan Levi}, \bibinfo{person}{Liran Gispan}, \bibinfo{person}{Niv Giladi}, {and} \bibinfo{person}{Ethan Fetaya}.} \bibinfo{year}{2022}\natexlab{}.
\newblock \showarticletitle{Evaluating and calibrating uncertainty prediction in regression tasks}.
\newblock \bibinfo{journal}{\emph{Sensors}} \bibinfo{volume}{22}, \bibinfo{number}{15} (\bibinfo{year}{2022}), \bibinfo{pages}{5540}.
\newblock


\bibitem[Ng et~al\mbox{.}(2002)]%
        {ng2002algorithm}
\bibfield{author}{\bibinfo{person}{Kwok-suen Ng}, \bibinfo{person}{Wing-tat Hung}, {and} \bibinfo{person}{Wing-gun Wong}.} \bibinfo{year}{2002}\natexlab{}.
\newblock \showarticletitle{An algorithm for assessing the risk of traffic accident}.
\newblock \bibinfo{journal}{\emph{Journal of safety research}} \bibinfo{volume}{33}, \bibinfo{number}{3} (\bibinfo{year}{2002}), \bibinfo{pages}{387--410}.
\newblock


\bibitem[Niculescu-Mizil and Caruana(2005)]%
        {niculescu2005predicting}
\bibfield{author}{\bibinfo{person}{Alexandru Niculescu-Mizil} {and} \bibinfo{person}{Rich Caruana}.} \bibinfo{year}{2005}\natexlab{}.
\newblock \showarticletitle{Predicting good probabilities with supervised learning}. In \bibinfo{booktitle}{\emph{Proceedings of the 22nd international conference on Machine learning}}. \bibinfo{pages}{625--632}.
\newblock


\bibitem[Nixon et~al\mbox{.}(2019)]%
        {nixon2019measuring}
\bibfield{author}{\bibinfo{person}{Jeremy Nixon}, \bibinfo{person}{Michael~W Dusenberry}, \bibinfo{person}{Linchuan Zhang}, \bibinfo{person}{Ghassen Jerfel}, {and} \bibinfo{person}{Dustin Tran}.} \bibinfo{year}{2019}\natexlab{}.
\newblock \showarticletitle{Measuring Calibration in Deep Learning.}. In \bibinfo{booktitle}{\emph{CVPR workshops}}, Vol.~\bibinfo{volume}{2}.
\newblock


\bibitem[Pearce et~al\mbox{.}(2018)]%
        {pearce2018high}
\bibfield{author}{\bibinfo{person}{Tim Pearce}, \bibinfo{person}{Alexandra Brintrup}, \bibinfo{person}{Mohamed Zaki}, {and} \bibinfo{person}{Andy Neely}.} \bibinfo{year}{2018}\natexlab{}.
\newblock \showarticletitle{High-quality prediction intervals for deep learning: A distribution-free, ensembled approach}. In \bibinfo{booktitle}{\emph{International conference on machine learning}}. PMLR, \bibinfo{pages}{4075--4084}.
\newblock


\bibitem[Qian et~al\mbox{.}(2023)]%
        {qian2023uncertainty}
\bibfield{author}{\bibinfo{person}{Weizhu Qian}, \bibinfo{person}{Dalin Zhang}, \bibinfo{person}{Yan Zhao}, \bibinfo{person}{Kai Zheng}, {and} \bibinfo{person}{JQ James}.} \bibinfo{year}{2023}\natexlab{}.
\newblock \showarticletitle{Uncertainty quantification for traffic forecasting: A unified approach}. In \bibinfo{booktitle}{\emph{2023 IEEE 39th International Conference on Data Engineering (ICDE)}}. IEEE, \bibinfo{pages}{992--1004}.
\newblock


\bibitem[Rahman(2023)]%
        {rahman_uncertainty-aware_2023}
\bibfield{author}{\bibinfo{person}{Moshiur Rahman}.} \bibinfo{year}{2023}\natexlab{}.
\newblock \showarticletitle{Uncertainty-{Aware} {Traffic} {Prediction} using {Attention}-based {Deep} {Hybrid} {Network} with {Bayesian} {Inference}}.
\newblock \bibinfo{journal}{\emph{International Journal of Advanced Computer Science and Applications}} \bibinfo{volume}{14}, \bibinfo{number}{6} (\bibinfo{year}{2023}).
\newblock


\bibitem[Rudin(2019)]%
        {rudin2019stop}
\bibfield{author}{\bibinfo{person}{Cynthia Rudin}.} \bibinfo{year}{2019}\natexlab{}.
\newblock \showarticletitle{Stop explaining black box machine learning models for high stakes decisions and use interpretable models instead}.
\newblock \bibinfo{journal}{\emph{Nature machine intelligence}} \bibinfo{volume}{1}, \bibinfo{number}{5} (\bibinfo{year}{2019}), \bibinfo{pages}{206--215}.
\newblock


\bibitem[Sankararaman and Mahadevan(2013)]%
        {sankararaman2013distribution}
\bibfield{author}{\bibinfo{person}{S Sankararaman} {and} \bibinfo{person}{S Mahadevan}.} \bibinfo{year}{2013}\natexlab{}.
\newblock \showarticletitle{Distribution type uncertainty due to sparse and imprecise data}.
\newblock \bibinfo{journal}{\emph{Mechanical Systems and Signal Processing}} \bibinfo{volume}{37}, \bibinfo{number}{1-2} (\bibinfo{year}{2013}), \bibinfo{pages}{182--198}.
\newblock


\bibitem[Shen et~al\mbox{.}(2023)]%
        {shen2023post}
\bibfield{author}{\bibinfo{person}{Maohao Shen}, \bibinfo{person}{Yuheng Bu}, \bibinfo{person}{Prasanna Sattigeri}, \bibinfo{person}{Soumya Ghosh}, \bibinfo{person}{Subhro Das}, {and} \bibinfo{person}{Gregory Wornell}.} \bibinfo{year}{2023}\natexlab{}.
\newblock \showarticletitle{Post-hoc Uncertainty Learning Using a Dirichlet Meta-Model}. In \bibinfo{booktitle}{\emph{Proceedings of the AAAI Conference on Artificial Intelligence}}, Vol.~\bibinfo{volume}{37}. \bibinfo{pages}{9772--9781}.
\newblock


\bibitem[Song et~al\mbox{.}(2019)]%
        {song2019distribution}
\bibfield{author}{\bibinfo{person}{Hao Song}, \bibinfo{person}{Tom Diethe}, \bibinfo{person}{Meelis Kull}, {and} \bibinfo{person}{Peter Flach}.} \bibinfo{year}{2019}\natexlab{}.
\newblock \showarticletitle{Distribution calibration for regression}. In \bibinfo{booktitle}{\emph{International Conference on Machine Learning}}. PMLR, \bibinfo{pages}{5897--5906}.
\newblock


\bibitem[Thiagarajan et~al\mbox{.}(2020)]%
        {thiagarajan2020building}
\bibfield{author}{\bibinfo{person}{Jayaraman~J Thiagarajan}, \bibinfo{person}{Bindya Venkatesh}, \bibinfo{person}{Prasanna Sattigeri}, {and} \bibinfo{person}{Peer-Timo Bremer}.} \bibinfo{year}{2020}\natexlab{}.
\newblock \showarticletitle{Building calibrated deep models via uncertainty matching with auxiliary interval predictors}. In \bibinfo{booktitle}{\emph{Proceedings of the AAAI Conference on Artificial Intelligence}}, Vol.~\bibinfo{volume}{34}. \bibinfo{pages}{6005--6012}.
\newblock


\bibitem[Vose(2008)]%
        {vose2008risk}
\bibfield{author}{\bibinfo{person}{David Vose}.} \bibinfo{year}{2008}\natexlab{}.
\newblock \bibinfo{booktitle}{\emph{Risk analysis: a quantitative guide}}.
\newblock \bibinfo{publisher}{John Wiley \& Sons}.
\newblock


\bibitem[Wang et~al\mbox{.}(2023)]%
        {wang2023uncertainty}
\bibfield{author}{\bibinfo{person}{Qingyi Wang}, \bibinfo{person}{Shenhao Wang}, \bibinfo{person}{Dingyi Zhuang}, \bibinfo{person}{Haris Koutsopoulos}, {and} \bibinfo{person}{Jinhua Zhao}.} \bibinfo{year}{2023}\natexlab{}.
\newblock \showarticletitle{Uncertainty Quantification of Spatiotemporal Travel Demand with Probabilistic Graph Neural Networks}.
\newblock \bibinfo{journal}{\emph{arXiv preprint arXiv:2303.04040}} (\bibinfo{year}{2023}).
\newblock


\bibitem[Wu et~al\mbox{.}(2021a)]%
        {wu2021quantifying}
\bibfield{author}{\bibinfo{person}{Dongxia Wu}, \bibinfo{person}{Liyao Gao}, \bibinfo{person}{Matteo Chinazzi}, \bibinfo{person}{Xinyue Xiong}, \bibinfo{person}{Alessandro Vespignani}, \bibinfo{person}{Yi-An Ma}, {and} \bibinfo{person}{Rose Yu}.} \bibinfo{year}{2021}\natexlab{a}.
\newblock \showarticletitle{Quantifying uncertainty in deep spatiotemporal forecasting}. In \bibinfo{booktitle}{\emph{Proceedings of the 27th ACM SIGKDD Conference on Knowledge Discovery \& Data Mining}}. \bibinfo{pages}{1841--1851}.
\newblock


\bibitem[Wu and James(2021)]%
        {wu2021bayesian}
\bibfield{author}{\bibinfo{person}{Ying Wu} {and} \bibinfo{person}{JQ James}.} \bibinfo{year}{2021}\natexlab{}.
\newblock \showarticletitle{A bayesian learning network for traffic speed forecasting with uncertainty quantification}. In \bibinfo{booktitle}{\emph{2021 International Joint Conference on Neural Networks (IJCNN)}}. IEEE, \bibinfo{pages}{1--7}.
\newblock


\bibitem[Wu et~al\mbox{.}(2023)]%
        {wu2023adaptive}
\bibfield{author}{\bibinfo{person}{Ying Wu}, \bibinfo{person}{Yongchao Ye}, \bibinfo{person}{Adnan Zeb}, \bibinfo{person}{James~JQ Yu}, {and} \bibinfo{person}{Zheng Wang}.} \bibinfo{year}{2023}\natexlab{}.
\newblock \showarticletitle{Adaptive Modeling of Uncertainties for Traffic Forecasting}.
\newblock \bibinfo{journal}{\emph{arXiv preprint arXiv:2303.09273}} (\bibinfo{year}{2023}).
\newblock


\bibitem[Wu et~al\mbox{.}(2021b)]%
        {wu2021inductive}
\bibfield{author}{\bibinfo{person}{Yuankai Wu}, \bibinfo{person}{Dingyi Zhuang}, \bibinfo{person}{Aurelie Labbe}, {and} \bibinfo{person}{Lijun Sun}.} \bibinfo{year}{2021}\natexlab{b}.
\newblock \showarticletitle{Inductive graph neural networks for spatiotemporal kriging}. In \bibinfo{booktitle}{\emph{Proceedings of the AAAI Conference on Artificial Intelligence}}, Vol.~\bibinfo{volume}{35}. \bibinfo{pages}{4478--4485}.
\newblock


\bibitem[Wu et~al\mbox{.}(2019)]%
        {wu2019graph}
\bibfield{author}{\bibinfo{person}{Zonghan Wu}, \bibinfo{person}{Shirui Pan}, \bibinfo{person}{Guodong Long}, \bibinfo{person}{Jing Jiang}, {and} \bibinfo{person}{Chengqi Zhang}.} \bibinfo{year}{2019}\natexlab{}.
\newblock \showarticletitle{Graph wavenet for deep spatial-temporal graph modeling}.
\newblock \bibinfo{journal}{\emph{arXiv preprint arXiv:1906.00121}} (\bibinfo{year}{2019}).
\newblock


\bibitem[Yu et~al\mbox{.}(2017)]%
        {yu2017spatio}
\bibfield{author}{\bibinfo{person}{Bing Yu}, \bibinfo{person}{Haoteng Yin}, {and} \bibinfo{person}{Zhanxing Zhu}.} \bibinfo{year}{2017}\natexlab{}.
\newblock \showarticletitle{Spatio-temporal graph convolutional networks: A deep learning framework for traffic forecasting}.
\newblock \bibinfo{journal}{\emph{arXiv preprint arXiv:1709.04875}} (\bibinfo{year}{2017}).
\newblock


\bibitem[Zadrozny and Elkan(2001)]%
        {zadrozny2001obtaining}
\bibfield{author}{\bibinfo{person}{Bianca Zadrozny} {and} \bibinfo{person}{Charles Elkan}.} \bibinfo{year}{2001}\natexlab{}.
\newblock \showarticletitle{Obtaining calibrated probability estimates from decision trees and naive bayesian classifiers}. In \bibinfo{booktitle}{\emph{Icml}}, Vol.~\bibinfo{volume}{1}. \bibinfo{pages}{609--616}.
\newblock


\bibitem[Zhao et~al\mbox{.}(2022)]%
        {zhao2022multi}
\bibfield{author}{\bibinfo{person}{Xiangyu Zhao}, \bibinfo{person}{Wenqi Fan}, \bibinfo{person}{Hui Liu}, {and} \bibinfo{person}{Jiliang Tang}.} \bibinfo{year}{2022}\natexlab{}.
\newblock \showarticletitle{Multi-type urban crime prediction}. In \bibinfo{booktitle}{\emph{Proceedings of the AAAI Conference on Artificial Intelligence}}, Vol.~\bibinfo{volume}{36}. \bibinfo{pages}{4388--4396}.
\newblock


\bibitem[Zheng et~al\mbox{.}(2018)]%
        {zheng2018hybrid}
\bibfield{author}{\bibinfo{person}{Yao Zheng}, \bibinfo{person}{Qianqian Zhu}, \bibinfo{person}{Guodong Li}, {and} \bibinfo{person}{Zhijie Xiao}.} \bibinfo{year}{2018}\natexlab{}.
\newblock \showarticletitle{Hybrid quantile regression estimation for time series models with conditional heteroscedasticity}.
\newblock \bibinfo{journal}{\emph{Journal of the Royal Statistical Society Series B: Statistical Methodology}} \bibinfo{volume}{80}, \bibinfo{number}{5} (\bibinfo{year}{2018}), \bibinfo{pages}{975--993}.
\newblock


\bibitem[Zhou et~al\mbox{.}(2021)]%
        {zhou2021stuanet}
\bibfield{author}{\bibinfo{person}{Zhengyang Zhou}, \bibinfo{person}{Yang Wang}, \bibinfo{person}{Xike Xie}, \bibinfo{person}{Lei Qiao}, {and} \bibinfo{person}{Yuantao Li}.} \bibinfo{year}{2021}\natexlab{}.
\newblock \showarticletitle{STUaNet: Understanding uncertainty in spatiotemporal collective human mobility}. In \bibinfo{booktitle}{\emph{Proceedings of the Web Conference 2021}}. \bibinfo{pages}{1868--1879}.
\newblock


\bibitem[Zhuang et~al\mbox{.}(2022)]%
        {zhuang2022uncertainty}
\bibfield{author}{\bibinfo{person}{Dingyi Zhuang}, \bibinfo{person}{Shenhao Wang}, \bibinfo{person}{Haris Koutsopoulos}, {and} \bibinfo{person}{Jinhua Zhao}.} \bibinfo{year}{2022}\natexlab{}.
\newblock \showarticletitle{Uncertainty quantification of sparse travel demand prediction with spatial-temporal graph neural networks}. In \bibinfo{booktitle}{\emph{Proceedings of the 28th ACM SIGKDD Conference on Knowledge Discovery and Data Mining}}. \bibinfo{pages}{4639--4647}.
\newblock


\bibitem[Zuo et~al\mbox{.}(2020)]%
        {zuo2020transformer}
\bibfield{author}{\bibinfo{person}{Simiao Zuo}, \bibinfo{person}{Haoming Jiang}, \bibinfo{person}{Zichong Li}, \bibinfo{person}{Tuo Zhao}, {and} \bibinfo{person}{Hongyuan Zha}.} \bibinfo{year}{2020}\natexlab{}.
\newblock \showarticletitle{Transformer hawkes process}. In \bibinfo{booktitle}{\emph{International conference on machine learning}}. PMLR, \bibinfo{pages}{11692--11702}.
\newblock


\end{thebibliography}

\clearpage
\appendix
\section{Appendix}
\subsection{Modified Model Implementation Details}
\label{sec:modify_model_appendix}
To implement the modified GNNs, we use the Github repositories for STGCN \footnote{\url{https://github.com/FelixOpolka/STGCN-PyTorch}}, GWN \footnote{\url{https://github.com/nnzhan/Graph-WaveNet}}, and DSTAGNN \footnote{\url{https://github.com/SYLan2019/DSTAGNN}} respectively. 
We basically keep all the hyper-parameters and important parameters as the default from the original repositories, which can be referred to in our Github repository \footnote{\url{https://github.com/AnonymousSAUC/SAUC}}.
Notice that we keep the same model parameters in both the original models and the modified models, which can be found in Table \ref{tab:appendix_params}

\begin{table}[htbp]
    \centering
    \begin{tabular}{cccc}
    \toprule
       Parameters/Models  & STGCN & GWN & DSTAGNN \\
    \midrule
       Input window length  & 12 & 12 & 12 \\
       Output window length  & 12 & 12 & 12 \\
       Learning rate & 0.001 & 0.001 & 0.001\\
       Batch size & 24 & 24 & 24 \\
       Output layer & Linear & Conv2d & Linear\\
       Training epochs & 1000 & 100 & 500\\
       Early stop& No & Yes & Yes \\
    \bottomrule
    \end{tabular}
    \caption{Parameters of original and modified GNN models.}
    \label{tab:appendix_params}
\end{table}

\begin{figure}[!ht]
    \centering    
    \subfigure[Training Loss of STGCN on CCR dataset]{\includegraphics[scale=0.4]{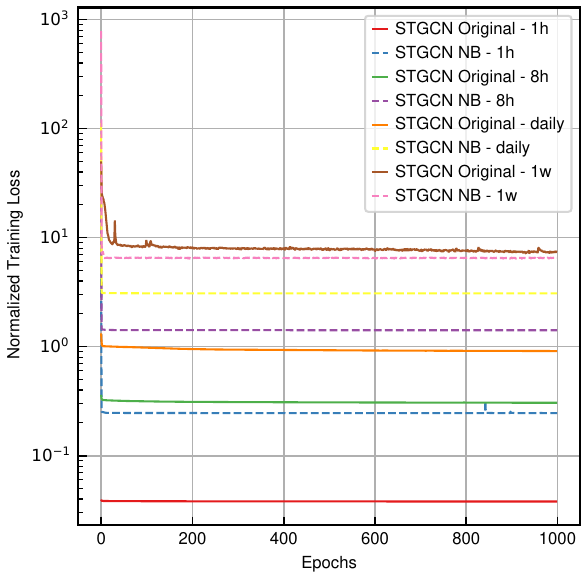}}
    \subfigure[Training Loss of GWN on CCR dataset]{\includegraphics[scale=0.4]{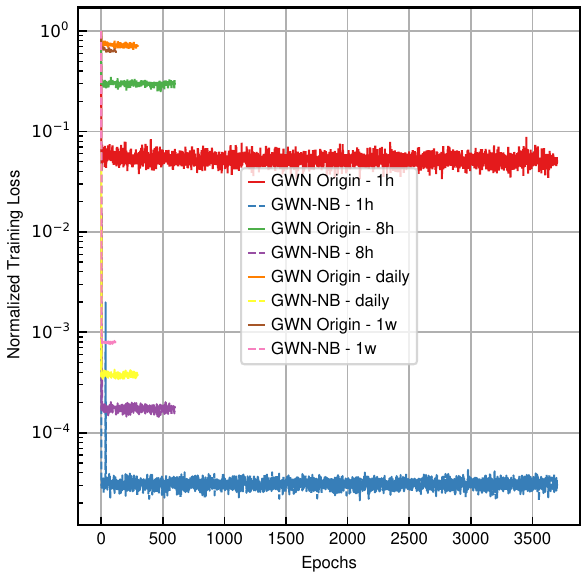}}
    \caption{Training loss of STGCN, GWN, and their modifications on the CCR dataset.}
    \label{fig:appendix_training_loss_ccr}
\end{figure}

Training loss on CCR datasets can be found in Figure \ref{fig:appendix_training_loss_ccr}. All models are converged properly.

The full implementation results can be found in Figure \ref{tab:appendix_modification}. 
It is clear that the model's performance
in terms of numerical accuracy is close to the results using models before modification. 
In fact, the NB modification shows better performance in the sparse dataset as the NB distributions are more suitable for the nature of discrete data. 

\begin{table}[ht]
    \small
    \centering
    \resizebox{\linewidth}{!}{%
    \begin{tabular}{cccccccc}
    \toprule
        Dataset & Error Metric & STGCN & STGCN-NB & GWN & GWN-NB & DSTAGNN & DSTAGNN-NB \\        
    \midrule
        \multirow{3}{*}{CCR\_1h} 
        & MAE  & 0.429 & 0.317 & 0.848 & 0.334 & 0.982 & 0.901\\
        & RMSE & 0.637 & 0.703 & 0.962 & 0.766 & 1.226 & 1.117 \\
    \midrule
        \multirow{3}{*}{CCR\_8h} 
        & MAE  & 1.499 & 1.452 & 1.395 & 1.478 & 1.475 & 1.198\\
        & RMSE & 2.386 & 2.326 & 2.052 & 2.376 & 2.272 & 2.693 \\
    \midrule
        \multirow{3}{*}{CCR\_daily} 
        & MAE  & 2.951 & 2.719 & 2.521 & 2.681 & 2.500 & 2.421 \\
        & RMSE & 5.167 & 4.286 & 3.784 & 9.703 & 3.848 & 3.368 \\
    \midrule
        \multirow{3}{*}{CCR\_weekly} 
        & MAE  & 14.146 & 12.505 & 9.938 & 14.516 & 10.389 & 17.341\\
        & RMSE & 23.333 & 20.242 & 16.103 & 23.231 & 16.578 & 21.382\\
    \midrule
        \multirow{3}{*}{CTC\_1h} 
        & MAE  & 0.084 & 0.044 & 0.960 & 0.044 & 1.049 & 0.832 \\
        & RMSE & 0.214 & 0.220 & 0.980 & 0.220 & 1.074 & 0.871 \\
    \midrule
        \multirow{3}{*}{CTC\_8h} 
        & MAE  & 0.453 & 0.346 & 0.810 & 0.352 & 1.113 & 0.995 \\
        & RMSE & 0.633 & 0.726 & 0.924 & 0.746 & 1.289 & 1.177 \\
    \midrule
        \multirow{3}{*}{CTC\_daily} 
        & MAE  & 0.861 & 1.433 & 0.895 & 1.019 & 1.299 & 0.986 \\
        & RMSE & 1.148 & 1.786 & 1.154 & 1.550 & 1.596 & 1.353 \\
    \midrule
        \multirow{3}{*}{CTC\_weekly} 
        & MAE  & 2.474 & 3.618 & 2.423 & 3.279 & 2.502 & 2.956\\
        & RMSE & 3.317 & 3.929 & 3.178 & 3.689 & 3.256 & 3.876 \\
    \bottomrule
    \end{tabular}%
    }
    \caption{Comparisons between modified models and the original models.}
    \label{tab:appendix_modification}
\end{table}

\subsection{Baseline Model Implementation}
\label{sec:baseline_implement_appendix}
We list the introductions and pseudo codes to implement the baseline models.
\subsubsection{Temperature Scaling}
Temperature scaling is a parametric method that modifies the outputs of a regression model. This modification is controlled by a single parameter, the temperature \( T \). The algorithms can be formulated as follows:

\begin{algorithm}[ht]
\caption{Temperature Scaling}
\begin{algorithmic}[1]
\STATE Train a model on a training dataset, yielding point predictions \( f(x) \).
\STATE Optimize the temperature \( T \) on the validation set by minimizing the mean squared error between the predicted and true values:
\[ T = \arg \min_T \frac{1}{n} \sum_{i=1}^n (f(x_i) - y_i)^2 \]
\FOR{each new input \( x \)}
\STATE Compute the model's prediction \( f(x) \).
\STATE Apply temperature scaling with the learned \( T \) to the prediction \( f(x) \), yielding scaled prediction:
\[ g(f(x)) = \frac{f(x)}{T} \]
\ENDFOR
\end{algorithmic}
\end{algorithm}

Temperature scaling is an efficient and straightforward method for calibration that requires optimization of only a single parameter. By scaling the outputs, it can yield calibrated predictions without modifying the rank ordering of the model's predictions.

\subsubsection{Isotonic Regression}
Isotonic regression is a non-parametric method utilized for the calibration of a predictive model's outputs. In the context of regression, this method operates as shown below:

\begin{algorithm}[ht]
\caption{Isotonic Regression for Regression}
\begin{algorithmic}[1]
\STATE Train a model on a training dataset, providing point predictions \( f(x) \).
\STATE Fit an isotonic regression model on the validation set:
\STATE \quad Sort the predictions \( f(x_i) \) and corresponding true values \( y_i \) in ascending order of \( f(x_i) \).
\STATE \quad Solve the isotonic regression problem to find a non-decreasing sequence \( g(f(x)) \) that minimizes:
\[ \min \sum_{i=1}^n (g(f(x_i)) - y_i)^2 \]
\FOR{each new input \( x \)}
\STATE Compute the model's prediction \( f(x) \).
\STATE Use the fitted isotonic regression model to convert the prediction $f(x)$ to a calibrated prediction: $g(f(x))$
\ENDFOR
\end{algorithmic}
\end{algorithm}

Isotonic regression makes no assumptions about the form of the function connecting the model's predictions to calibrated predictions. This flexibility allows it to fit complex, non-linear mappings, which can provide improved calibration performance when the model's outputs are not well-modeled by a simple function.

\subsubsection{Histogram Binning}
Histogram binning can also be applied in a regression setting to calibrate predictions. The idea is to split the range of your model's outputs into several bins, and then adjust the predictions within each bin to match the average actual outcome within that bin. The specific algorithm is:

\begin{algorithm}[ht]
\caption{Histogram Binning}
\begin{algorithmic}[1]
\STATE Train a model on a training dataset, yielding predictions \( f(x) \).
\STATE Determine the bins for the predictions on the validation set.
\FOR{each bin \( B_i \)}
\STATE Compute the average true value \( \bar{y}_i \) for all examples in the bin:
\[ \bar{y}_i = \frac{1}{|B_i|} \sum_{x_j \in B_i} y_j \]
\STATE Adjust the prediction for each example in the bin to the average true value \( \bar{y}_i \).
\ENDFOR
\FOR{each new input \( x \)}
\STATE Compute the model's prediction \( f(x) \).
\STATE Find the bin \( B_i \) that \( f(x) \) falls into and adjust \( f(x) \) to the average true value for that bin:
\[ g(f(x)) = \bar{y}_i \]
\ENDFOR
\end{algorithmic}
\end{algorithm}

This method assumes that predictions within each bin are uniformly mis-calibrated. It is less flexible than isotonic regression, but it is simpler and less prone to overfitting, especially when the number of bins is small.

\subsubsection{Platt Scaling}
In a regression setting, Platt Scaling can be interpreted as applying a sigmoid function transformation to the model's outputs. The output is then considered as the mean of a Bernoulli distribution. While this might be useful in some contexts, it is not generally applicable to regression problems but can be useful in sparse datasets. A linear transformation is fitted to the model's predictions to minimize the mean squared error on the validation set. The parameters of this transformation are learned from the data, making this an instance of post-hoc calibration. Note that this adaptation might not be suitable for all regression tasks, especially those where the target variable has a non-linear relationship with the features.

\begin{algorithm}[ht]
\caption{Platt Scaling}
\begin{algorithmic}[1]
\STATE Train a model on a training dataset, yielding predictions $f(x)$.
\STATE Minimize the mean squared error on the validation set between $g(f(x))$ and the true outcomes, where $g(f(x)) = af(x) + b$ is a linear transformation of the model's predictions:
$ a, b = \arg \min_{a, b} \frac{1}{n} \sum_{i=1}^n (af(x_i) + b - y_i)^2$
\FOR{each new input  $x$ }
\STATE Compute the model's prediction $f(x)$.
\STATE Apply the learned linear transformation to $f(x)$ to get the calibrated prediction: $g(f(x)) = af(x) + b$
\ENDFOR
\end{algorithmic}
\end{algorithm}

\end{document}